%% file: main.tex
\documentclass{article}

\usepackage{microtype}
\usepackage{graphicx}
\usepackage{subfigure}
\usepackage{booktabs} % for professional tables

\usepackage{hyperref}

\usepackage[accepted]{mlsys2022}

\usepackage[utf8]{inputenc} % allow utf-8 input
\usepackage[T1]{fontenc}    % use 8-bit T1 fonts
\usepackage{hyperref}       % hyperlinks
\usepackage{url}            % simple URL typesetting
\usepackage{booktabs}       % professional-quality tables
\usepackage{amsfonts}       % blackboard math symbols
\usepackage{nicefrac}       % compact symbols for 1/2, etc.
\usepackage{microtype}      % microtypography
\usepackage[bottom]{footmisc} % prevent figure drop below footnote

\usepackage{times}
\usepackage{graphicx}
\usepackage[ruled,vlined,linesnumbered]{algorithm2e}
\graphicspath{{figure/}}

\usepackage[margin=0.12cm]{caption}
\usepackage{multirow}
\usepackage{etoolbox}

\usepackage{enumitem}

\usepackage{wrapfig}
\usepackage{array}
\newcommand{\PreserveBackslash}[1]{\let\temp=\\#1\let\\=\temp}
\newcolumntype{C}[1]{>{\PreserveBackslash\centering}p{#1}}
\newcolumntype{R}[1]{>{\PreserveBackslash\raggedleft}p{#1}}
\newcolumntype{L}[1]{>{\PreserveBackslash\raggedright}p{#1}}

\usepackage{microtype}
\usepackage{graphicx}
\usepackage{subfigure}
\usepackage{booktabs} % for professional tables
\usepackage{amsmath}
\usepackage{amsfonts}
\usepackage{amsthm}
\usepackage{float}

\newtheorem{assumption}{Assumption}[section]

\newcommand{\expe}{\mathbb{E}}

\SetKwComment{Comment}{$\triangleright$\ }{}

\SetKwFunction{FMain}{Main}

\SetCommentSty{mycommfont}

\newcommand{\niparagraph}[1]{\noindent\textbf{#1}}

\mlsystitlerunning{BNS-GCN: Efficient Full-Graph Training of Graph Convolutional Networks with Boundary Node Sampling}

\begin{document}

\twocolumn[
\mlsystitle{BNS-GCN: Efficient Full-Graph Training of Graph Convolutional Networks with Partition-Parallelism and Random Boundary Node Sampling}

\mlsyssetsymbol{equal}{*}

\begin{mlsysauthorlist}
\mlsysauthor{Cheng Wan$^*$}{rice}
\mlsysauthor{Youjie Li$^*$}{uiuc}
\mlsysauthor{Ang Li}{pnnl}
\mlsysauthor{Nam Sung Kim}{uiuc}
\mlsysauthor{Yingyan Lin}{rice}
\end{mlsysauthorlist}

\mlsysaffiliation{rice}{Rice University}
\mlsysaffiliation{uiuc}{University of Illinois at Urbana-Champaign}
\mlsysaffiliation{pnnl}{Pacific Northwest National Laboratory}

\mlsyscorrespondingauthor{Yingyan Lin}{yingyan.lin@rice.edu}

\mlsyskeywords{Graph Neural Networks, Graph Convolutional Networks, Full-Graph Training, Large-Graph Training, Distributed Training, Partition Parallelism, Sampling, Machine Learning}

\vskip 0.3in

\begin{abstract}

Graph Convolutional Networks (GCNs) have emerged as the state-of-the-art method for graph-based learning tasks.
However, training GCNs at scale is still challenging, hindering both the exploration of more sophisticated GCN architectures and their applications to real-world large graphs.
While it might be natural to consider graph partition and distributed training for tackling this challenge, this direction has only been slightly scratched the surface in the previous works due to the limitations of existing designs.
In this work, we first analyze why distributed GCN training is ineffective and identify the underlying cause to be the excessive number of boundary nodes of each partitioned subgraph, which easily explodes the memory and communication costs for GCN training. 
Furthermore, we propose a simple yet effective method dubbed BNS-GCN that adopts random Boundary-Node-Sampling to enable efficient and scalable distributed GCN training. 
Experiments and ablation studies consistently validate the effectiveness of BNS-GCN, e.g., boosting the throughput by up to 16.2$\times$ and reducing the memory usage by up to 58\%, while maintaining a full-graph accuracy.
Furthermore, both theoretical and empirical analysis show that BNS-GCN enjoys a better convergence than existing sampling-based methods. 
We believe that our BNS-GCN has opened up a new paradigm for enabling GCN training at scale. 
The code is available at \href{https://github.com/RICE-EIC/BNS-GCN}{https://github.com/RICE-EIC/BNS-GCN}.

\end{abstract}
]

\printAffiliationsAndNotice{\mlsysEqualContribution} % otherwise use the standard text.

\input{section/introduction}

\input{section/background}

\input{section/method}

\input{section/experiment}
\input{section/conclusion}

\section*{ACKNOWLEDGEMENT}
The work is supported by the National Science Foundation (NSF) through the MLWiNS program (Award number: 2003137), the CC$^*$ Compute program (Award number: 2019007) and the NeTS program (Award number: 1801865).

\bibliography{bns}
\bibliographystyle{mlsys2022}

\appendix

\input{section/appendix}

\end{document}

%% file: section/introduction.tex
\section{Introduction}

Graph convolutional networks (GCNs) \citep{kipf2016semi} have emerged as the state-of-the-art (SOTA) method for various graph-based learning tasks, including node classification \citep{kipf2016semi}, link prediction \citep{zhang2018link}, graph classification \citep{xu2018powerful}, and recommendation systems \citep{ying2018graph}.
The outstanding performance of GCNs is attributed to their unrestricted and irregular neighborhood connectivity, which provides them a greater applicability to graph-based data than convolutional neural networks (CNNs) that adopt a fixed regular neighborhood structure.
Specifically, given a node in a graph, a GCN first \textit{aggregates} the features of its neighbors and then \textit{updates} its own feature through a hierarchical feed-forward propagation. 
The two dominant operations, \textit{aggregate} and \textit{update} of node features, enables GCNs to take advantage of the graph structure and thus outperform their structure-unaware alternatives.

Despite their promising performance, training GCNs at scale has been very challenging, thereby hindering the exploration of more sophisticated GCN architectures and restricting their real-world applications to large graphs.
This is because as the graph size grows, the sheer number of node features and the giant adjacency matrix can easily explode the required memory and communications.
To tackle this challenge, several sampling-based methods have been developed at a cost of approximation errors. 
For example, GraphSAGE \citep{hamilton2017inductive} and VR-GCN \citep{chen2018stochastic} reduce a full graph into a mini-batch via neighbor sampling; alternative methods \citep{chiang2019cluster,zeng2019graphsaint} extract sub-graphs as training samples.

In parallel with sampling-based methods, a more recent direction for handling large-graph training is distributed GCN training, which aims at training a large \textit{full-graph} over multiple GPUs without degrading the accuracy. 
The key idea is to partition a giant graph into small subgraphs such that each can fit a single GPU, and train them in parallel with necessary communication.
Following this ``partition-parallelism'' paradigm, pioneering efforts (NeuGraph \citep{ma2019neugraph}, ROC \citep{jia2020improving}, CAGNET \citep{tripathy2020reducing}, Dorylus \citep{thorpe2021dorylus}, and PipeGCN \citep{wan2022pipegcn}) have demonstrated a promising training performance.
Nonetheless, these works still suffer from heavy communication traffics, 
limiting their achievable training efficiency, let alone the potentially harmful staleness due to asynchronous training~\citep{thorpe2021dorylus}.

To enable scalable and efficient large-graph GCN training without compromising the full-graph accuracy, this work sets out to understand the underlying cause of the communication and memory explosion in distributed GCNs training 
and finds that distributed GCN training can be ineffective if it is not designed properly, 
which motivates us to make the following contributions:
\begin{itemize}

   \item We first analyze and identify three main challenges in partition-parallel training of GCNs: (1) overwhelming communication volume, (2) prohibitive memory requirement, and (3) imbalanced memory consumption. 
   We further localize their cause to be an excessive number of \textit{boundary nodes} (rather than boundary edges) associated with each partitioned subgraph, which is unique to GCNs due to their \textit{neighbor aggregation} (see Section~\ref{sec:BNS-cha}).
	This finding enhances the understanding in distributed GCN training and can potentially inspire further ideas in this direction. 
	
	\item To tackle all above challenges in one shot, we propose a simple yet effective method dubbed BNS-GCN which 
	randomly samples features of \textit{boundary nodes} at each training iteration and achieves a \textbf{triple win} -- aggressively shrinking the communication and memory requirements while leading to a better generalization accuracy (see Section~\ref{sec:BNS-GCN}). 
	\textit{To the best of our knowledge, this is the first work directly targeting at reducing the communication volume in distributed GCN training, without incurring extra computing resource overhead (e.g., CPU) or hurting the achieved accuracy}.

	\item We further provide theoretical analysis to validate the improved convergence offered by BNS-GCN (see Section~\ref{sec:BNS-theory}). 
	Extensive experiments and ablation studies consistently validate the benefit of BNS-GCN in both training efficiency and accuracy, e.g., boosting the throughput by up to 16.2$\times$ and reducing the memory usage by up to 58\% while achieving the same or an even better accuracy, over the SOTA methods, when being applied to Reddit, ogbn-products, Yelp, and ogbn-papers100M datasets (see Section~\ref{sec:experiment}).
\end{itemize}

%% file: section/background.tex
\section{Background and Related Works}
\niparagraph{Graph Convolutional Networks.}
GCNs take graph-structured data as inputs and learn feature vectors (embedding) for each node of a graph.
Specifically,
GCN performs two major steps in each layer, i.e., \textit{neighbor aggregation} and \textit{node update}, which can be represented as:
\begin{equation}
z^{(\ell)}_v=\zeta^{(\ell)}\left(\left\{h^{(\ell-1)}_u\mid u\in\mathcal{N}(v)\right\}\right)\label{fml:aggr}
\end{equation}
\begin{equation}
h^{(\ell)}_v=\phi^{(\ell)}\left(z^{(\ell)}_v,h^{(\ell-1)}_v\right)\label{fml:update}
\end{equation}
where $\mathcal{N}(v)$ denotes the neighbor set of node $v$ in the graph, $h^{(\ell)}_u$ denotes the learned feature vector of node $u$ at the $\ell$-th layer, 
$\zeta^{(\ell)}$ denotes the aggregation function that takes neighbor features to generate aggregation result $z_v^{(\ell)}$ for node $v$, and finally $\phi^{(\ell)}$ gets the feature of node $v$ updated. 
A famous instance of GCNs is GraphSAGE with a mean aggregator~\citep{hamilton2017inductive}, in which $\zeta^{(\ell)}$ is the mean function and $\phi^{(\ell)}$ is $ \sigma\left(W^{(\ell)}\cdot\textup{CONCAT}\left(z^{(\ell)}_v,h^{(\ell-1)}_v\right)\right)$, where $W^{(\ell)}$ is the weight matrix and $\sigma$ is a non-linear activation. While we mainly use this instance for evaluating our BNS-GCN, our approach can be easily extended to other popular aggregators and update functions.
\begin{figure*}[!t]
    \begin{center}
    \includegraphics[width=1\linewidth]{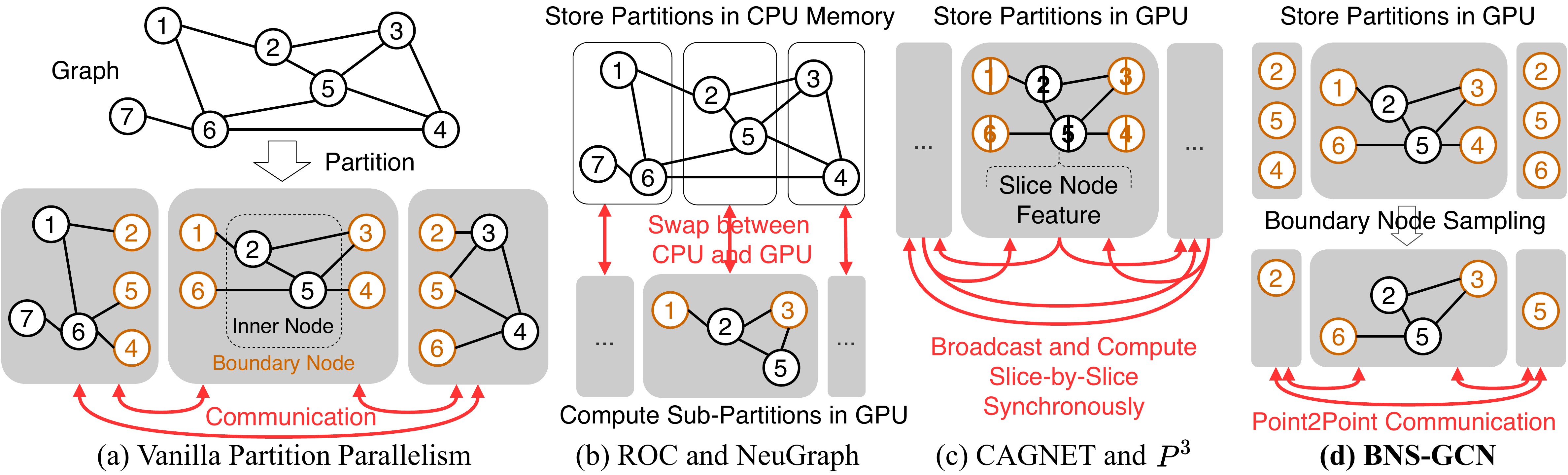}
    \end{center}
    \caption{An illustrative comparison between existing distributed GCN training methods and our BNS-GCN.}
    \label{fig:ConceptCompare}
\end{figure*}

\niparagraph{Sampling-Based GCN Training.}
Real-world graphs consist of millions of nodes and 
edges \citep{hu2020open}, far beyond the capability of vanilla GCNs. 
As such, sampling-based methods were proposed, e.g., neighbor sampling \citep{hamilton2017inductive,chen2018stochastic}, layer sampling \citep{chen2018fastgcn,huang2018adaptive,zou2019layer}, and subgraph sampling \citep{chiang2019cluster,zeng2019graphsaint}, which yet suffer from:
\begin{itemize}
\item \textit{Inaccurate feature estimation}: although most sampling methods provide unbiased estimation of node features, the variance of these estimation hurts the model accuracy. As \citep{cong2020minimal} shows, a smaller variance is beneficial to improving the accuracy of a sampling-based method;
\item \textit{Neighbor explosion}: \citet{hamilton2017inductive} first uses node sampling to randomly select several neighbors in the previous layer, but as GCNs get deeper the size of selected nodes exponentially increases. \citet{chen2018stochastic} further proposes samplers for restricting the size of neighbor expansion, which yet suffers from heavy memory requirements;
\item \textit{Sampling overhead}: All sampling-based methods incur extra time for generating mini-batches, which can occupy 25\%+ of the training time \citep{zeng2019graphsaint}.
\end{itemize}

\niparagraph{Distributed Training for GCNs.}
To train GCNs for real-world large graphs, distributed training leveraging multiple GPUs to enable full-graph training has been shown to be promising.  
Nevertheless, GCNs training is different from the challenge of classical distributed DNN training where (1) data samples are small yet the model is large (model parallelism \citep{krizhevsky2014one,harlap2018pipedream}) and (2) data samples do not have dependency (data parallelism \citep{li2020pytorch,li2018pipe,li2018inceptionn}), both violating the nature of GCNs. As such,
GCN-oriented methods should partition the full graph into small subgraphs such that each could be fitted into a single GPU memory, and train them in parallel, where communication across subgraphs is necessary to exchange boundary node features to perform GCNs' neighbor aggregation, which is called \textit{vanilla partition parallelism} as shown in Figure~\ref{fig:ConceptCompare}(a).
Following this paradigm, several works have been proposed.
ROC \citep{jia2020improving}, NeuGraph \citep{ma2019neugraph}, and AliGraph \citep{zhu2019aligraph} partition large graphs and store all partitions in CPUs and swaps a fraction of each partition to compute in GPUs (see Figure~\ref{fig:ConceptCompare}(b)). 
Their training efficiency are thus compromised due to expensive CPU-GPU swaps.
CAGNET \citep{tripathy2020reducing} and $P^3$ \citep{gandhi2021p3} further split node features and layers to enable intra-layer model parallelism (see Figure~\ref{fig:ConceptCompare}(c)), which however incurs a heavy communication overhead especially when the feature dimension is large.
Dorylus \citep{thorpe2021dorylus} improves the vanilla partition parallelism by pipelining each fine-grain computation operation in GCN training over numerous CPU threads, which still suffers from the communication bottleneck.

\niparagraph{Distributed Graph Systems.}
Distributed graph systems were proposed to solve general graph problems~\citep{gonzalez2012powergraph,shun2013ligra,nguyen2013lightweight,zhu2016gemini,chen2019powerlyra}. 
\citep{lerer2019pytorch} also proposes a distributed learning system for graph embedding. 
However, none of these considers node features and hence cannot be used for GCN training.

%% file: section/method.tex
\begin{figure*}[t]
\begin{center}
\includegraphics[width=1.0\linewidth]{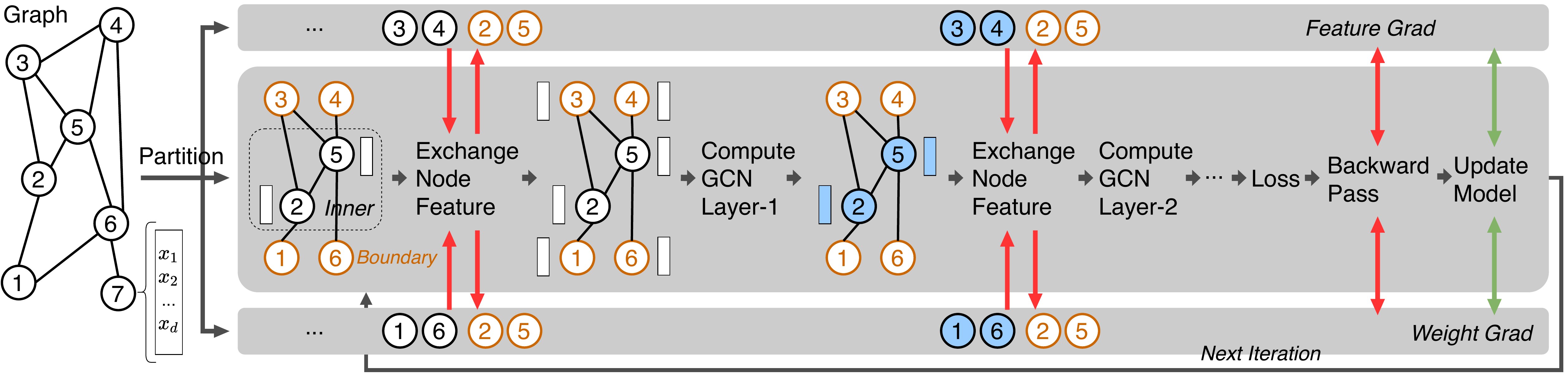}
\end{center}
\caption{Illustrating vanilla partition parallel training. A large graph is partitioned into smaller subgraphs (see the inner nodes in black) with each being able to fit into one GPU memory. The key challenge is that excessive boundary nodes (in orange) associated with each subgraph (due to GCNs' neighbor aggregation) can lead to a heavy communication overhead, extra memory cost, and memory imbalance among subgraphs, thus limiting the achievable scalability and efficiency of distributed GCN training.}
\label{fig:framework}
\end{figure*}

\begin{table*}[t]
\centering
\caption{Comparison between the number of \textit{boundary} nodes and \textit{inner} nodes in partitioned Reddit graph~\citep{hamilton2017inductive}. The standard METIS~\citep{karypis1998fast} is used for graph partition.}
\label{tab:partition}
\setlength{\tabcolsep}{0.8em}
\begin{tabular}{rrrrrrrrrrr}
\hline
Partition Index                  & 1    & 2    & 3    & 4    & 5    & 6    & 7    & 8    & 9    & 10   \\
\hline 
\# Inner Nodes                   & 14k  & 15k  & 15k  & 15k  & 15k  & 15k  & 14k  & 15k  & 14k  & 15k  \\
\# Boundary Nodes                & 39k  & 15k  & 86k  & 78k  & 86k  & 62k  & 6k   & 46k  & 71k  & 23k  \\
Ratio of \# Boundary to \# Inner & \textbf{2.64} & 1.00 & \textbf{5.45} & \textbf{4.95} & \textbf{5.49} & \textbf{4.11} & 0.42 & \textbf{3.04} & \textbf{4.81} & 1.52 \\
\hline
\end{tabular}
\end{table*}

\section{The Proposed BNS-GCN Framework}
\label{sec:method}
\niparagraph{Overview.} 
To address all aforementioned limitations (see Figure~\ref{fig:ConceptCompare}(a-c)), we propose partition-parallel training of GCNs with \textbf{B}oundary \textbf{N}ode \textbf{S}ampling, dubbed \textbf{BNS-GCN}, as shown in Figure~\ref{fig:ConceptCompare}(d).
BNS-GCN partitions a full-graph with minimized boundary nodes and then further randomly samples the boundary nodes to shrink both communication and memory costs, enabling efficient large-graph training while maintaining the full-graph accuracy.
We develop BNS-GCN by first analyzing the three major challenges in partition-parallel training of GCNs and then pinpoint their underlying cause (see Section~\ref{sec:BNS-cha}).
To tackle the cause directly, we design a simple yet effective sampling strategy that can \textit{simultaneously alleviate all three challenges} (see Section~\ref{sec:BNS-GCN}), while achieving \textit{much reduced} variances (i.e., closer to that of the full-graph one) of feature approximation as compared to existing sampling methods (see Section~\ref{sec:BNS-theory}). 
We further discuss the difference between BNS-GCN and other sampling-based methods (see Section~\ref{sec:BNS-Discuss}) for better understanding our new contribution.
\subsection{Challenges in Partition-Parallel Training}
\label{sec:BNS-cha}
\label{sec:vanilla}
To enable full-graph training, the original graph can be partitioned into smaller subgraphs (i.e., partitions) to be trained locally on each accelerator/node while communicating dependent node features across subgraphs, which is termed as \textit{partition parallelism}. 
As shown in Figure~\ref{fig:framework}, each subgraph contains a subset of nodes from the original graph, termed as an \textit{inner node set} (see ``Inner'').
Additionally, each subgraph holds a \textit{boundary node set} (see ``Boundary'') containing dependent nodes from other subgraphs. 
Such a \textit{boundary node set} is dictated by GCNs' neighbor aggregation from neighbor subgraphs, e.g., node-5 in Figure~\ref{fig:framework} requires nodes-[3,4,6] residing on other subgraphs to perform Equation~\ref{fml:aggr}, creating the \textit{boundary nodes} associated with the subgraph hosting node-5. 
To compute each GCN layer, features of boundary nodes are communicated or exchanged across subgraphs (shown in red) before those of inner nodes get updated (e.g., nodes-[2,5] in blue). 
The updated features are again exchanged across subgraphs to compute the next GCN layer, which is repeated until the final layer. 
The backward pass follows a similar process but communicates gradients of boundary nodes instead of features. 
Afterwards, the GCN model gets updated via weight gradient sharing (in green) among partitions using AllReduce.

However, vanilla partition parallelism is neither inefficient nor scalable due to the following three major challenges:
\begin{enumerate}[label=\roman*]
    \item \textit{\textbf{Heavy Communication Overhead}} is resulting from exchanging boundary nodes across partitions, limiting the scalability to larger graphs or using more partitions.
   
    \item \textit{\textbf{Prohibitive Memory Requirement}} is incurred in each partition to hold both the inner and boundary sets, the latter of which can overflow a GPU's memory capacity.
    
    \item \textit{\textbf{Imbalanced Memory Requirements}} exists across all partitions, where the memory straggler (i.e., the partition requiring a significantly larger memory than others) not only determines the memory requirement but also causes under-utilization of other partitions' GPUs.
  
\end{enumerate}

We identify that all three challenges above share the same underlying cause -- \textit{the overhead of extra \textbf{boundary nodes} associated with each partition due to distributed partitions}.

\niparagraph{Communication Cost Analysis of Vanilla Partition Parallelism.}
For a partition $\mathcal{G}_i$, its communication volume can be defined as $\text{Vol}(\mathcal{G}_i)=\sum_{v\in \mathcal{G}_i}D(v)$ where $D(v)$ is the number of different partitions in which $v$ has at least one neighbor node, excluding $\mathcal{G}_i$ \citep{bulucc2016recent}. 
This value quantifies the total amount of features $\mathcal{G}_i$ needs to send during each propagation (Equation~\ref{fml:aggr}). 
As the total number of received messages equals to the total number of sent messages, the \textit{total communication volume equals to the total number of boundary nodes (instead of boundary edges)}:
\begin{equation}
\text{Vol}_\text{total}=\sum_i\text{Vol}(\mathcal{G}_i)=\sum_in_{bd}^{(i)}\label{fml:comm}
\end{equation}
where $n_{bd}^{(i)}$ is the number of boundary nodes in partition $\mathcal{G}_i$.

\niparagraph{Memory Cost Analysis of Vanilla Partition Parallelism.}
For a $\ell$-th layer, suppose the input feature is of dimension $d^{(\ell)}$, and the numbers of inner nodes and boundary nodes in partition $\mathcal{G}_i$ are $n_{in}^{(i)}$ and $n_{bd}^{(i)}$, respectively.
Considering a general case where all node features and inner nodes' aggregated features are saved for the back propagation in both Equation~\ref{fml:aggr} and Equation~\ref{fml:update}. 
When using a GraphSAGE layer with a mean aggregator, the memory cost is:
\begin{equation}
\text{Mem}^{(\ell)}(\mathcal{G}_i)=(3n_{in}^{(i)}+n_{bd}^{(i)})d^{(\ell)}\label{fml:mem}
\end{equation}
As a result, \textit{the memory requirement increases linearly with the number of boundary nodes (instead of boundary edges)}.

The challenge is that \textit{the number of boundary nodes can be excessive}.
Table~\ref{tab:partition} shows a typical example, where
the number of boundary nodes in each partition can be as high as $5.5\times$ of that of inner nodes, leading to both prohibitive communication and memory overhead. 

\begin{figure}[t]
    \centering
    \includegraphics[width=1.0\linewidth]{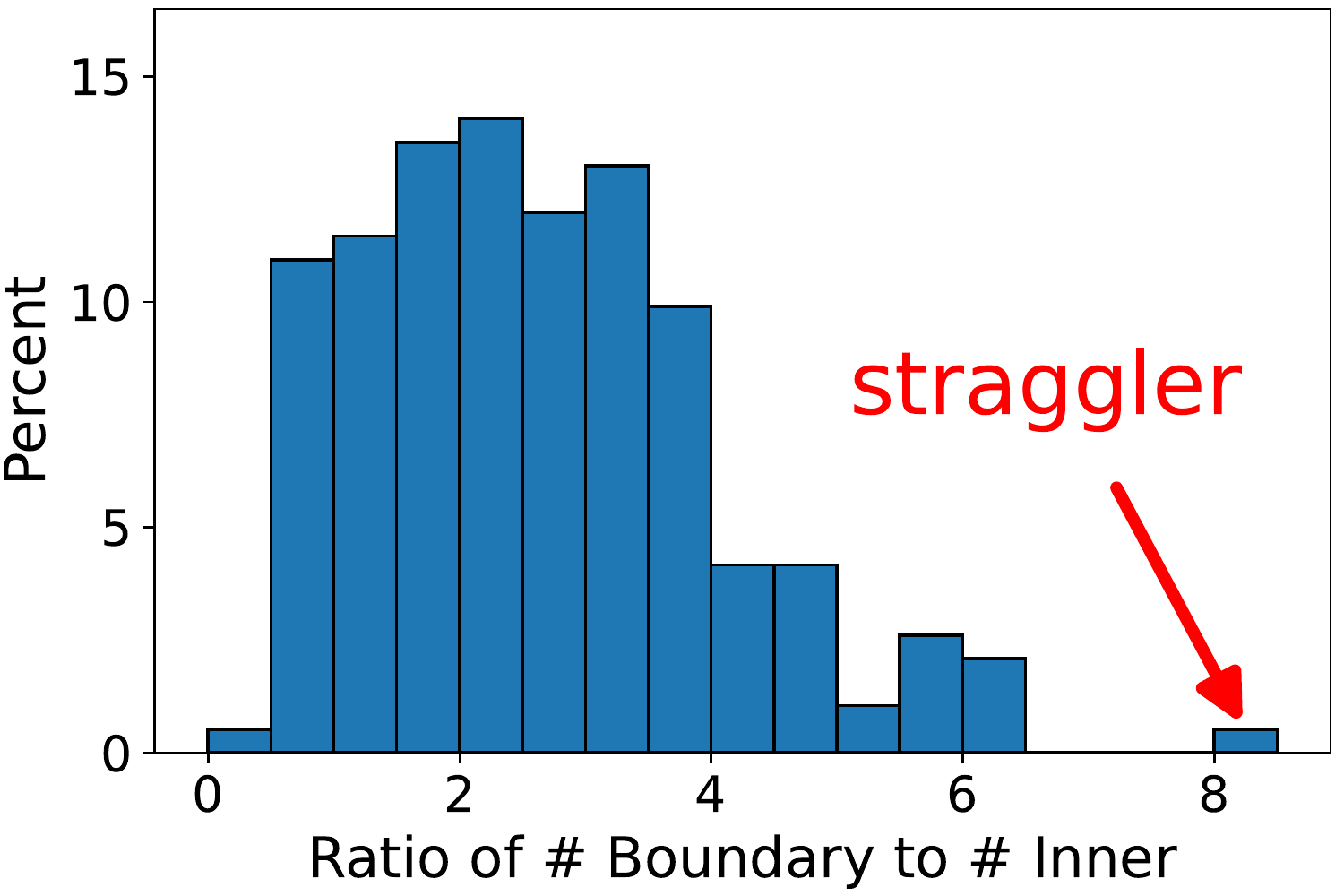}
        \caption{The distributions of the boundary-inner ratios for the ogbn-papers100M dataset under 192 partitions.}
    \label{fig:bd_dist}
\end{figure}

\niparagraph{Memory Imbalance Analysis in Vanilla Partition Parallelism.}
The memory cost can be highly imbalanced across partitions due to the irregular amounts of boundary nodes despite the balanced amount of inner nodes (see Table~\ref{tab:partition}). 
Furthermore, when scaling up to more partitions, the memory imbalance becomes more severe.
Figure~\ref{fig:bd_dist} shows such an example where we split a giant graph (ogbn-papers100M~\citep{hu2020open}) into 192 parts. 
The memory straggler (the one with boundary-inner ratio of 8) costs significantly more memory than other partitions, which not only raises the memory requirement but also incurs memory under-utilization for all other partitions' GPUs.
\subsection{The Proposed BNS-GCN Technique}

\label{sec:BNS-GCN}
\niparagraph{Graph Partition.}
As boundary nodes are the cause for the efficiency bottleneck of partition parallelism, the graph partition has to \textit{minimize all boundary node sets} to minimize subsequent communication and memory overheads, dubbed Goal-1. 
Besides, the graph partition must also achieve \textit{balanced computation time} across all partitions, dubbed Goal-2, since partition parallelism is a synchronous training paradigm that requires frequent synchronization at each layer (again due to GCNs' neighbor aggregation), under which unbalanced partition results in stragglers that block other partitions to proceed.

Prior works~\citep{tripathy2020reducing,zheng2020distdgl} aim at achieving only Goal-2 yet ignore Goal-1, while this work achieves both.
For Goal-2, we approximate the computational complexity of each node, aiming to balance computations across all partitions (e.g., when GraphSAGE computation is dominated by Equation~\ref{fml:update}, the complexity is proportional to the number of nodes, so we set partitions with an equal size in this case).
Then we optimize the graph partition algorithm for Goal-1.
In this work, the popular METIS~\citep{karypis1998fast} is adopted as the default graph partition algorithm and its objective is set to minimize the communication volume, i.e., minimize the number of boundary nodes (Equation~\ref{fml:comm}).
Besides METIS, other partitioning algorithms are also compatible with BNS-GCN (see Tables~\ref{tab:rand_part}-\ref{tab:rand_comm}). 
Note that the time complexity of METIS is $\mathcal{O}(|\mathcal{E}|)$ where $\mathcal{E}$ is the set of edges, and only needs to be performed once during the preprocessing stage, the cost of which can thus be amortized over numerous training iterations and leads to a negligible overhead. 
In addition, METIS is widely adopted in scalable GCN training~\citep{zhu2019aligraph,zheng2020distdgl,fey2021gnnautoscale,wan2022pipegcn} where the objective function is mostly set as the minimum cut (i.e., minimize the number of edges).

{
\makeatletter
\newcommand{\removelatexerror}{\let\@latex@error\@gobble}
\makeatother
\begin{figure}[!t]
\removelatexerror
\SetAlCapHSkip{0em}
\makeatletter
\patchcmd{\@algocf@start}% <cmd>
  {-1.5em}% <search>
  {0pt}% <replace>
  {}{}% <success><failure>
\makeatother

\begin{minipage}{0.477\textwidth}
\centering
\begin{algorithm}[H]
\SetAlgoLined
\KwIn{partition number $m$, partition id $i$, graph partition $\mathcal{G}_i$, boundary node set $\mathcal{B}_i$, node feature $X_i$, label $Y_i$, sampling rate $p$, initial model $w[0]$, learning rate $\eta$}
\KwOut{trained model $w[T]$ after $T$ iterations}
$\mathcal{V}_i\leftarrow \{\text{node }v\in\mathcal{G}_i:v\notin\mathcal{B}_i\}$\Comment*[r]{create inner node set}
$H^{(0)}\leftarrow X_i$\Comment*[r]{initialize input features}
\For{$t\leftarrow1:T$}{
  $\mathcal{U}_i \leftarrow$ randomly pick elements in $\mathcal{B}_i$ with probability $p$\label{line:select}\;
  
  $\mathcal{H}_i \leftarrow$ node induced subgraph of $\mathcal{G}_i$ from $\mathcal{V}_i\cup\mathcal{U}_i$\label{line:subgraph}\;
  
  Broadcast $\mathcal{U}_i$ and Receive $[\mathcal{U}_1,\cdots,\mathcal{U}_m]$\label{line:comm_idx}\;
  
  $[\mathcal{S}_{i,1},\cdots,\mathcal{S}_{i,m}]\leftarrow[\mathcal{U}_1\cap\mathcal{V}_i,\cdots,\mathcal{U}_m\cap\mathcal{V}_i$]\label{line:record}\;
  
  \For{$\ell\leftarrow1:L$}{
  	Send $[H_{\mathcal{S}_{i,1}}^{(\ell-1)},\cdots,H_{\mathcal{S}_{i,m}}^{(\ell-1)}]$ to partition $[1,\cdots,m]$ and Receive $H_{\mathcal{U}_i}^{(\ell-1)}$\label{line:comm_feat}\;
  	
  	$H^{(\ell)}\leftarrow GCN^{(\ell)}\left(\mathcal{H}_i,\left[\begin{array}{c}H^{(\ell-1)}\\ H_{\mathcal{U}_i}^{(\ell-1)}\end{array}\right], w[t-1]\right)$\label{line:comp}\;
  }
  $f_i\leftarrow\sum_{v\in\mathcal{V}_i}loss(h_v^{(L)}, y_v)$\Comment*[r]{calculate loss}
  $g_i[t]\leftarrow \frac{\partial f_i}{\partial w[t-1]}$\Comment*[r]{backward pass\label{line:backward}}
  $g[t]\leftarrow AllReduce (g_i[t])$\Comment*[r]{share gradients\label{line:allreduce}}
  $w[t]\leftarrow w[t-1]-\eta\cdot g[t]$\Comment*[r]{update model\label{line:update}}
}
\Return $w[T]$
\caption{Boundary node sampling for partition-parallel training (per-partition view)}
\label{alg:bs}
\end{algorithm}
\end{minipage}
\end{figure}
}

\niparagraph{Boundary Node Sampling (BNS).}
Even with an optimal graph partition, the boundary node issue still remains (see Table~\ref{tab:partition}), calling for innovative methods to reduce the boundary node volume.
An ideal method should achieve three goals: 
(1) substantially shrinking the size of boundary node sets, 
(2) incurring a minimal overhead, 
and (3) maintaining the full-graph accuracy. 
As such, we adopt a random sampling method called \textit{boundary node sampling}.
The key idea is to independently select a subset of boundary nodes from each partition, then \textbf{\textit{to store and communicate merely those selected ones instead of the full boundary sets}}, with a random selection varying from one epoch to another.

\begin{table}[t]
\setlength{\tabcolsep}{0.1em}
\caption{Comparing feature approximation variance between SOTA sampling methods and BNS-GCN, where we fix the target node set $\mathcal{V}_i$ across all methods. Here $\gamma$ denotes the upper bound of the $L_2$-norm of intermediate features, and $\Delta\gamma$ is the upper bound of the difference between the embedding feature and its history. We report the variance by ignoring the same factors. Note that $\left|\mathcal{B}_i\right| \ll \left|\mathcal{N}_i\right|\ll\left|\mathcal{V}\right|$.}
\label{tab:variance}
\begin{tabular}{@{}ccl@{}}
\hline
Method & Variance & Notation \\
\hline
GraphSAGE & $\mathcal{O}(D\gamma^2/s_{n})$ & \multirow{2}{4.05cm}{$s_n$: sampled neighbor size\\$D$: average degree}\\
VR-GCN & $\mathcal{O}(D\Delta\gamma^2/s_{n})$ & \\
\hline
FastGCN & $\mathcal{O}(|\mathcal{V}|\gamma^2/s_{\ell})$ & \multirow{3}{4.05cm}{$s_\ell$: sampled node set size\\$\mathcal{V},\mathcal{N}_i,\mathcal{B}_i$: global node set, neighbor set, boundary set} \\
LADIES &  $\mathcal{O}(|\mathcal{N}_i|\gamma^2/s_{\ell})$ & \\
\textbf{BNS-GCN} & $\mathcal{O}(|\mathcal{B}_i|\gamma^2/s_{\ell})$ &  \\
\hline 
\end{tabular}
\end{table}

Algorithm~\ref{alg:bs} outlines our proposed BNS-GCN. 
In the $i$-th partition, we randomly keep the boundary node set $\mathcal{U}_i$ with a probability $p$ and drop the rest at the beginning of each epoch (Lines~\ref{line:select}-\ref{line:subgraph}). 
These selected nodes' indices are then broadcasted among partitions such that each partition ``knows'' others' selections (Line~\ref{line:comm_idx}) and can also record its local node $\mathcal{S}_{i,j}$ that is selected by the other $j$-th partition (Line~\ref{line:record}). 
During the \textit{forward pass} of the $\ell$-th layer, each partition sends those features $H^{(\ell-1)}_{\mathcal{S}_{i,j}}$  of the previously recorded nodes to the corresponding $j$-th partition and meanwhile receives features $H^{(\ell-1)}_{\mathcal{U}_i}$ of its own selected boundary nodes to perform GCN operations (Lines~\ref{line:comm_feat}-\ref{line:comp}). 
For a mean aggregator, we replace the sent/received feature matrix $H$ with $H/p$ towards an unbiased feature estimation.
During the \textit{backward pass} of every layer, each partition sends and receives feature gradients of the selected boundary nodes while generating GCNs' weight gradients (Line~\ref{line:backward}).
Lastly, weight gradients are shared across partitions via AllReduce~\citep{thakur2005mpich} to perform weight updates (Lines~\ref{line:allreduce}-\ref{line:update}).

The proposed BNS-GCN reduces the number of boundary nodes by a factor of $\frac{1}{p}$, 
achieving a proportional reduction in both memory and communication costs (Equation~\ref{fml:comm}-\ref{fml:mem}).
Meanwhile, BNS-GCN pays negligible overhead due to its simplicity, which \textit{costs 0\%$\sim$7\% of the training time in practice}\footnote{Details can be found in Appendix~\ref{sec:sampling_overhead}.}.
Note that our BNS-GCN can not only boost the efficiency and scalability of vanilla partition parallelism, but also be easily plugged into any partition-parallel training methods (e.g., ROC and CAGNET) for further improving their training efficiency.

\subsection{Variance Analysis}
\label{sec:BNS-theory}
Theoretically, we study the effect of BNS-GCN on GCNs' performance by analyzing its feature approximation variance and comparing it with the SOTA methods. 
As the feature approximation variance controls the upper bound of gradient noise \citep{cong2020minimal}, a sampling method with a lower approximation variance usually enjoys a better convergence speed \citep{gower2019sgd} and higher accuracy.
Table~\ref{tab:variance} summarizes our results,  
where $\mathcal{V}$, $\mathcal{N}_i$, and $\mathcal{B}_i$ denote the global node set, neighbor set, and boundary neighbor set, respectively.
The detailed variance analysis of BNS-GCN can be found in Appendix~\ref{sec:var_proof} and variances for the other methods are based on \citep{zou2019layer}.
We find that \textit{\textbf{BNS-GCN enjoys the smallest variance}} compared with FastGCN and LADIES, when fixing the number of sampled nodes, as we strictly have $\mathcal{B}_i\subseteq\mathcal{N}_i\subseteq\mathcal{V}$.
To be able to compare with GraphSAGE, we fix the sampling size ($s_\ell$ = $s_n$), 
then BNS-GCN is strictly better than GraphSAGE, as BNS-GCN neither samples neighbors within the inner node sets nor samples the same nodes for multiple times, leading to $|\mathcal{B}_i|\leq D|\mathcal{V}_i|$.  
For VR-GCN, it is not comparable with BNS-GCN because the variance of VR-GCN is based on the difference between embedding feature and its history.

\subsection{BNS-GCN vs. Existing Sampling Methods}
\label{sec:BNS-Discuss}
We further discuss the difference between
BNS-GCN and existing sampling methods:

\begin{itemize}
    \item  \textit{Node Sampling}: GraphSAGE \citep{hamilton2017inductive} and VR-GCN \citep{chen2018stochastic} adopt node sampling which is likely to sample the same nodes multiple times from the previous layers, limiting GCNs' depth and training efficiency. 
Additionally, BNS-GCN does not sample neighbors within each subgragh, reducing both the estimation variance and sampling overhead.

\item \textit{Layer Sampling}: BNS-GCN is similar to layer sampling in that nodes within the same partition share the same sampled boundary nodes in the previous layer. 
Unlike FastGCN \citep{chen2018fastgcn}, AS-GCN \citep{huang2018adaptive} or LADIES \citep{zou2019layer}, BNS-GCN has much denser sampled layers, potentially leading to a higher accuracy.

\item \textit{Subgraph Sampling}: BNS-GCN could be viewed as one kind of subgraph sampling that drops boundary nodes from other partitions.
ClusterGCN \citep{chiang2019cluster} and GraphSAINT \citep{zeng2019graphsaint} propose subgraph sampling, yet their number of selected nodes are small, i.e., only 1.3\% and 5.3\% of the total nodes, respectively, causing a higher variance of gradient estimation.

\item \textit{Edge Sampling}: Applying edge sampling (e.g., DropEdge~\citep{rong2019dropedge}) to distributed GCN training is not efficient, as it does not directly reduce the number of boundary nodes (see Section~\ref{sec:compare_edge_sampling}). 
\end{itemize}

\begin{table*}[t]
\centering
\caption{Details of the graph datasets.}
\label{tab:setups}
\begin{tabular}{c|cccccc}
\hline
Dataset & \# Nodes & \# Edges & \# Feat. & \# Classes & Type & Train / Val / Test\\ \hline
Reddit~\citep{hamilton2017inductive} & 233K & 114M & 602 & 41 & inductive & 0.66 / 0.10 / 0.24 \\
ogbn-products~\citep{hu2020open}& 2.4M & 62M  & 100 & 47 & transductive & 0.08 / 0.02 / 0.90 \\
Yelp~\citep{zeng2019graphsaint} & 716K & 7.0M & 300 & 100 & inductive & 0.75 / 0.10 / 0.15 \\
ogbn-papers100M~\citep{hu2020open}& 111M & 1.6B & 128 & 172 & transductive & 0.78 / 0.08 / 0.14  \\ \hline
\end{tabular}
\end{table*}

%% file: section/experiment.tex
\begin{figure*}
  \centering
  \includegraphics[width=1.0\linewidth]{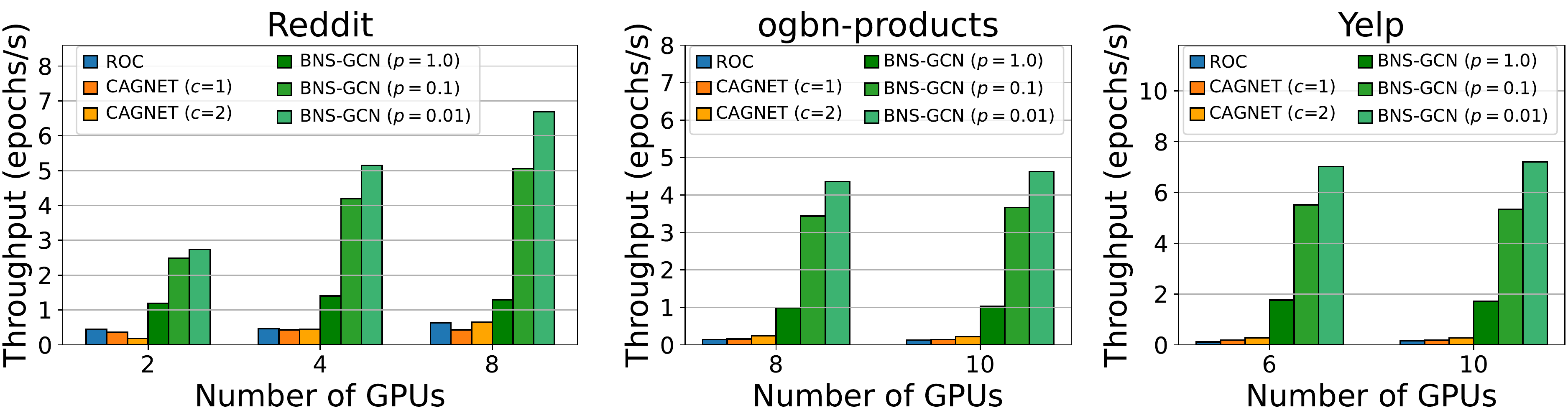}
  \caption{Throughput comparison on Reddit, ogbn-products, and Yelp. Each partition uses one GPU (except CAGNET ($c=2$) uses two). The boundary node sampling rate is denoted by $p$.}
  \label{fig:cmp}
\end{figure*}

\section{Experiments}
\label{sec:experiment}  

In this section, we first introduce our experiment setups, then compare with the SOTA baselines, and further provide ablation studies for a thorough evaluation on BNS-GCN.

\niparagraph{Datasets.} We evaluate BNS-GCN on four large-scale datasets:
1) Reddit~\citep{hamilton2017inductive} for community prediction based on the posts' contents and users' comments,
2) ogbn-products~\citep{hu2020open} for classifing Amazon products based on customers' review,
3) Yelp~\citep{zeng2019graphsaint} for predicting the types of business based on reviews and users' relationship,
and 4) ogbn-papers100M~\citep{hu2020open} for predicting the category of an arXiv publication based on its title and abstract.
Details of these four datasets are provided in Table~\ref{tab:setups}.

\niparagraph{Models.}
\label{sec:model}
We adopt a GraphSAGE model with an Adam optimizer for all datasets. The details are listed below:
\begin{itemize}
    \item Reddit: We use a 4-layer model with 256 hidden units and set the learning rate as 0.01 with 3000 epochs and 0.5 dropout rate.
    \item ogbn-products: We use a 3-layer model with 128 hidden units and set the learning rate as 0.003 with 500 epochs and 0.3 dropout rate.
    \item Yelp: We use a 4-layer model with 512 hidden units and set the learning rate as 0.001 with 3000 epochs and 0.1 dropout rate.
    \item ogbn-papers100M: We use a 3-layer model with 128 hidden units and set the learning rate as 0.01 with 100 epochs and 0.5 dropout rate.
\end{itemize}

\niparagraph{Setups.} 
We implement BNS-GCN in DGL~\citep{wang2019dgl} and PyTorch~\citep{paszke2019pytorch} with the default backend of Gloo. 
We conduct the experiments of Reddit,  ogbn-products and Yelp on a machine with 10 RTX-2080Ti (11GB), Xeon 6230R@2.10GHz (187GB), and PCIe3x16 connecting CPU-GPU and GPU-GPU. 
The minimal number of partitions for full-graph training are 2, 5, 3 for Reddit, ogbn-products, and Yelp, respectively. 
For ogbn-papers100M, the experiment is conducted on 32 machines, each of which has 6 Tesla V100 (16GB) with IBM Power9 (605GB).
To ensure the reproducibility and robustness of BNS-GCN, we do not tune but fix the hyper-parameters for BNS-GCN throughout all experiments, and we show evaluation results based on average of 10 runs.

\begin{table*}[t]
\centering
\caption{Comparison of test accuracy (\%) on Reddit and ogbn-products and of test F1-micro score (\%) on Yelp.}
\begin{tabular}{c|ccccccccc} 
\hline
Method  & \multicolumn{3}{c|}{Reddit} & \multicolumn{3}{c|}{ogbn-products}& \multicolumn{3}{c}{Yelp} \\ \hline
\multicolumn{10}{c}{{Sampling-based methods}} \\ \hline
FastGCN~\citep{chen2018fastgcn} & \multicolumn{3}{c|}{93.7} & \multicolumn{3}{c|}{60.42} & \multicolumn{3}{c}{26.5} \\
GraphSAGE~\citep{hamilton2017inductive} & \multicolumn{3}{c|}{95.4} & \multicolumn{3}{c|}{78.70} & \multicolumn{3}{c}{63.4} \\
\multicolumn{1}{c|}{AS-GCN~\citep{huang2018adaptive}} & \multicolumn{3}{c|}{96.3} & \multicolumn{3}{c|}{OOM$^\fnsymbol{footnote}$} & \multicolumn{3}{c}{OOM$^\fnsymbol{footnote}$} \\
\multicolumn{1}{c|}{LADIES~\citep{zou2019layer}} & \multicolumn{3}{c|}{94.3} & \multicolumn{3}{c|}{77.46} & \multicolumn{3}{c}{60.2} \\
\multicolumn{1}{c|}{VR-GCN~\citep{chen2018stochastic}} & \multicolumn{3}{c|}{96.3} & \multicolumn{3}{c|}{OOM$^\fnsymbol{footnote}$} & \multicolumn{3}{c}{64.0} \\
\multicolumn{1}{c|}{ClusterGCN~\citep{chiang2019cluster}} & \multicolumn{3}{c|}{96.6} & \multicolumn{3}{c|}{78.97} & \multicolumn{3}{c}{60.9} \\
\multicolumn{1}{c|}{GraphSAINT~\citep{zeng2019graphsaint}} & \multicolumn{3}{c|}{96.6} & \multicolumn{3}{c|}{79.08} & \multicolumn{3}{c}{65.3} \\ \hline
\multicolumn{10}{c}{\textbf{BNS-GCN}} \\ \hline
\multicolumn{1}{c|}{\# Partitions} & 2 & 4 & \multicolumn{1}{c|}{8} & 5 & 8 & \multicolumn{1}{c|}{10} & 3 & 6 & 10 \\ \hline
\multicolumn{1}{c|}{BNS-GCN ($p=1.0$)} & 97.11 & 97.11 & \multicolumn{1}{c|}{97.11} & 79.14 & 79.14 & \multicolumn{1}{c|}{79.14} & 65.26 & 65.26 & \multicolumn{1}{c}{65.26} \\
\multicolumn{1}{c|}{BNS-GCN ($p=0.1$)} & \textbf{97.17} & \textbf{97.16} & \multicolumn{1}{c|}{\textbf{97.15}} & 79.36 & \textbf{79.48} & \multicolumn{1}{c|}{\textbf{79.30}} & \textbf{65.32} & 65.26 & \textbf{65.34}\\
\multicolumn{1}{c|}{BNS-GCN ($p=0.01$)} & 97.09 & 96.99 & \multicolumn{1}{c|}{96.94} & \textbf{79.43} & 79.28 & \multicolumn{1}{c|}{79.21} & 65.27 & \textbf{65.31} & 65.29 \\
\multicolumn{1}{c|}{BNS-GCN ($p=0.0$)} & 97.03 & 96.88 & \multicolumn{1}{c|}{96.84} & 78.65 & 78.83 & \multicolumn{1}{c|}{78.79} & 65.28 & 65.27 & 65.23\\
\hline
\end{tabular}
\label{tab:acc}
\end{table*}

\subsection{
Comparison with the SOTA Baselines}
\niparagraph{Full-Graph Training Speedup.}
\label{sec:efficiency}
Figure~\ref{fig:cmp} compares the training throughput of BNS-GCN against the SOTA full-graph training methods, ROC\footnote{\href{https://github.com/jiazhihao/ROC}{https://github.com/jiazhihao/ROC}}~\citep{jia2020improving} and CAGNET\footnote{\href{https://github.com/PASSIONLab/CAGNET}{https://github.com/PASSIONLab/CAGNET}}~\citep{tripathy2020reducing}.
We observe that BNS-GCN consistently outperforms both baselines across different number of GPUs and boundary node sampling rates $p$. 
For the instance of training GCN on Reddit, BNS-GCN with $p=0.01$ offers a promising \textbf{\textit{throughput improvement of 8.9$\times$$\sim$16.2$\mathbf{\times}$}} over ROC and \textit{\textbf{9.2$\times$$\sim$13.8$\times$}} over CAGNET ($c=2$) across different number of GPUs.
Even when $p=1$, BNS-GCN still improves the throughput by 1.8$\times$$\sim$3.7$\times$ over ROC and 1.0$\times$$\sim$5.5$\times$ over CAGNET ($c=2$).
The advantage of BNS-GCN is attributed to not only the reduced communication overhead with \textit{boundary node sampling}, but also no swap between CPU and GPU as ROC nor redundant broadcast and synchronization overhead as CAGNET.
Furthermore, increasing the number of partitions boosts the performance of BNS-GCN ($p<1$) substantially, but not for other methods, validating 
BNS-GCN's advantageous scalability thanks to its effectiveness in reducing communication overhead by dropping \textit{boundary nodes}. 
The advantage of BNS-GCN is similar for the other two datasets.

\niparagraph{Full-Graph Accuracy.}
Now we show that BNS-GCN maintains or even improves the accuracy of full-graph training, while boosting the training efficiency.
Table~\ref{tab:acc} summarizes our extensive evaluations of test scores when BNS-GCN adopts various sampling rates and different numbers of partitions, and compare with seven SOTA sampling-based methods~\citep{hu2020open, redditacc2020paperwithcode,chiang2019cluster,chen2018stochastic,zeng2019graphsaint,hamilton2017inductive,cong2020minimal,liu2022exact,zou2019layer,chen2020scalable}.
We observe that \textit{full-graph training (BNS-GCN with $p=1$) always achieves a higher or comparable accuracy than existing sampling-based methods}, regardless of datasets or number of partitions, which is consistent with the results of ROC~\citep{jia2020improving}.
More importantly, \textit{BNS-GCN always maintains or even increases the full-graph accuracy}, regardless of the sampling rates (e.g., $p=0.1/0.01$), the number of partitions, or different datasets.
For instance, on Reddit, \textit{$p=0.1$ achieves a test accuracy of 97.15\%$\sim$97.17\% under 2$\sim$8 partitions, which are consistently better than the 97.11\% accuracy of full-graph unsampled training}, validating the effectiveness and robustness of BNS-GCN.
Meanwhile, we also observe that the \textit{special case of BNS-GCN}, $p=0$, always suffers from the worst test score on the three datasets, compared with other cases ($p>0$). We understand that this accuracy/score drop is due to the full isolation of each partition after completely removing all boundary nodes, leading to no boundary node features during neighbor aggregation throughout the end-to-end training.
To the best of our knowledge, \textbf{\textit{BNS-GCN achieves the best accuracy of training GraphSAGE-layer based GCNs on all three datasets compared with all existing works}}.

\begin{table}[t]
\centering
\caption{Comparison between BNS-GCN (10 partitions) and sampling-based methods on ogbn-products.}
\setlength{\tabcolsep}{0.4em}
\label{tab:cmp_samp}
\begin{tabular}{c|ccc}
\hline
Method & Total Train Time & Test Acc (\%) \\
\hline
ClusterGCN & 294.2s & 78.97±0.33 \\
NeighborSampling & 281.8s & 78.70±0.36 \\
GraphSAINT & 157.4s & 79.08±0.24 \\
\hline
BNS-GCN ($p=1.0$) & 269.1s & 79.14±0.35 \\
BNS-GCN ($p=0.1$) & 155.3s & \textbf{79.30±0.36} \\
BNS-GCN ($p=0.01$) & \textbf{142.9s} & 79.21±0.26 \\
\hline
\end{tabular}
\end{table}

\renewcommand{\thefootnote}{\fnsymbol{footnote}}
\footnotetext[1]{We find these cases run out of memory, which are consistent with~\citep{zeng2019graphsaint}.}
\renewcommand{\thefootnote}{\arabic{footnote}}

\begin{figure*}[t]
\centering
\begin{minipage}{.5\textwidth}
  \centering
  \includegraphics[width=0.93\linewidth]{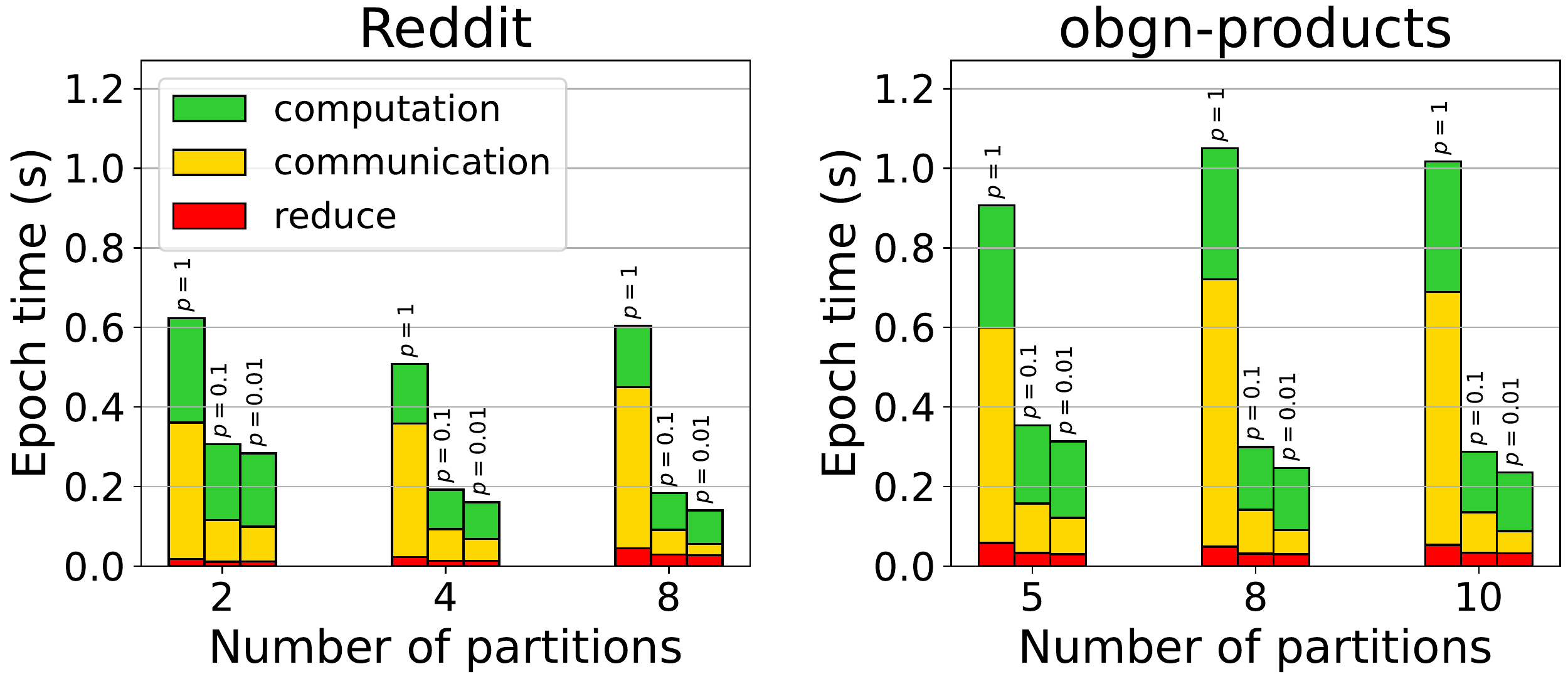}
  \caption{Training time breakdown of BNS-GCN with different boundary node sampling rates $p$.} 
  \label{fig:comm}
\end{minipage}%
\begin{minipage}{.5\textwidth}
  \centering
  \includegraphics[width=0.93\linewidth]{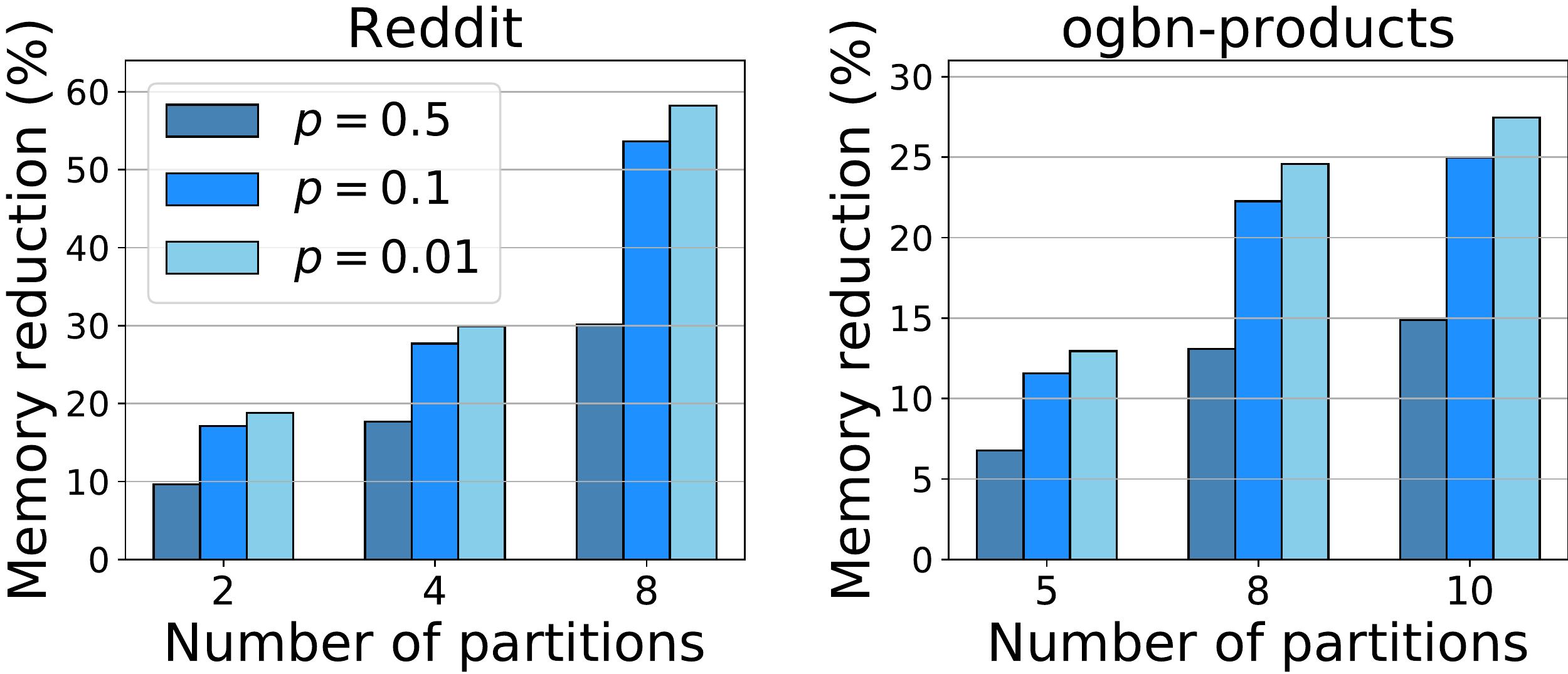}
  \caption{Memory usage reduction achieved by BNS-GCN, where the reduction is against $p=1.0$.}
  \label{fig:mem}
\end{minipage}
\end{figure*}

\begin{figure*}[t]
  \centering
  \includegraphics[width=1\linewidth]{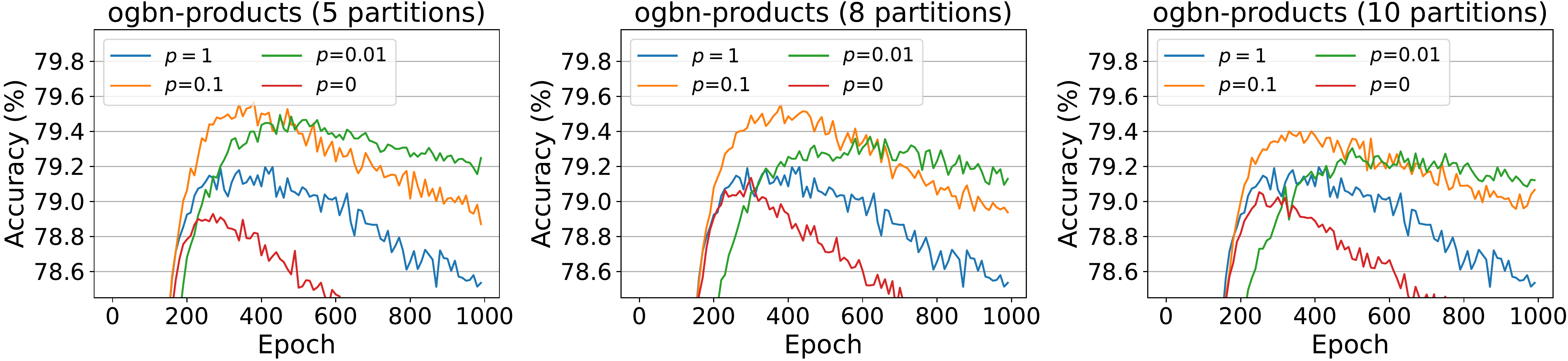}
  \caption{Test-accuracy convergence comparison among unsampled full-graph training (BNS-GCN with $p=1$), boundary-node sampled training ($p=0.1/0.01$), and isolated training ($p=0$) on ogbn-products.} 
  \label{fig:convergence_epoch}
\end{figure*}

\niparagraph{Improvement over Sampling-based Methods}.
Besides the full-graph training comparison, we also validate BNS-GCN's advantage over the SOTA sampling-based methods (implemented by the OGB team\footnote{\href{https://github.com/snap-stanford/ogb}{https://github.com/snap-stanford/ogb}} \citep{hu2020open}) on ogbn-products as shown in Table \ref{tab:cmp_samp}.
We observe that BNS-GCNs with $p=0.1$ and $p=0.01$ outperform all the sampling-based methods in terms of both efficiency and accuracy thanks to its lower approximation variance and substantially higher achieved throughput. 
More comparisons between BNS-GCN and sampling-based methods can be found in Appendix~\ref{sec:sampling_throughput}.

\subsection{Performance Analysis}

\niparagraph{Training Time Improvement Breakdown.} 
To further understand the improvement of BNS-GCN, we breakdown the training time into three major components (local computation, communication for boundary nodes, and allreduce on model gradient) as shown in Figure~\ref{fig:comm}.
We observe that \textit{\textbf{communication dominates the training time}} (up to 67\% and 64\% in baselines ($p=1$) on Reddit and obgn-products, respectively). 
As expected, with boundary node sampling ($p<1$), the communication overhead is substantially reduced, thus the total training time is improved. 
Specifically, $p=0.01$ sharply cuts \textbf{\textit{74\%$\sim$93\%}} and \textbf{\textit{83\%$\sim$91\%}} of the communication time from that of the baselines on Reddit and obgn-product, respectively, where this benefit consistently holds as the number of partitions scales.
Furthermore, in addition to single machine training, we also study BNS-GCN's benefits for multi-machine training and evaluate the performance on ogbn-papers100M. 
Specifically, 
we separate ogbn-papers100M into 192 parts and deploy the training on 32 machines (6 GPUs per machine), and provide results in Table~\ref{tab:papers100m}. We can see that
\textit{BNS-GCN with $p=0.01$ considerably reduce the total training time by \textbf{99\%}}, showing that distributed GCN training with multiple machines suffers from a more severe communication bottleneck and thus making BNS-GCN more desirable.

\begin{table}[t]
\centering
\caption{\textcolor{black}{Epoch time breakdown of ogbn-papers100M.
}}
\setlength{\tabcolsep}{0.3em}
\label{tab:papers100m}
\begin{tabular}{c|cccc}
\hline
Method & Total & Comp. & Comm. & Reduce \\ \hline
BNS-GCN ($p=1.0$) & 554.1s & 5.3s & 550.3s & 0.8s \\
BNS-GCN ($p=0.1$) & 58.7s & 1.0s & 56.9s & 0.8s \\
BNS-GCN ($p=0.01$) & 6.0s & 0.6s & 4.8s & 0.6s \\
\hline
\end{tabular}
\end{table}

\niparagraph{Memory Saving.} 
BNS-GCN's advantage in terms of memory usage reduction is shown in Figure~\ref{fig:mem}.
We observe that BNS-GCN consistently reduces the memory usage across different number of partitions on both graphs. 
Specifically, for the denser Reddit graph (with an average node degree of 984.0), $p=0.01$ saves 58\% memory usage for 8 GPUs. 
Even for the sparser obgn-products graph (with an average node degree of 50.5), $p=0.01$ still saves 27\% memory usage for 10 GPUs. 
Note that the memory saving of BNS-GCN scales with the numbers of partition, because the number of boundary nodes increases with more partitions, indicating BNS-GCN's scalability to training larger graphs.
Also, we find that BNS-GCN's memory reduction is not linear with reduced $p$, as besides the tensors analyzed in Equation~\ref{fml:mem} there are other objects (e.g., caches for non-linear activations and dropout) occupying the memory during training.

\begin{table*}[t]
\caption{Test score (\%) of BNS-GCN on top of random partition, where +/- shows the accuracy difference from BNS-GCN on top of METIS in Table~\ref{tab:acc}.}
\label{tab:rand_part}
\setlength{\tabcolsep}{0.8em}
\centering
\begin{tabular}{C{4cm}|C{1.1cm}C{1.1cm}|C{1.9cm}C{1.2cm}|C{1.2cm}C{1.2cm}}
\hline
Method & \multicolumn{2}{c|}{Reddit (8 partitions)} & \multicolumn{2}{c|}{ogbn-products (10 partitions)} & \multicolumn{2}{c}{Yelp (10 partitions)} \\ \hline
\textbf{Random}+BNS ($p=1.0$) & 97.11 & +0.00 & 79.14 & +0.00 & 65.26 & +0.00 \\
\textbf{Random}+BNS ($p=0.1$) & 96.95 & -0.20 & 79.57 & +0.27 & 65.18 & -0.16 \\
\textbf{Random}+BNS ($p=0.0$) & 93.37 & -3.47 & 75.39 & -3.40 & 64.92 & -0.31 \\ \hline
\end{tabular}
\end{table*}

\begin{table*}[t]
\caption{Training efficiency improvement of BNS-GCN ($p=0.1$) on top of different partition methods.}
\label{tab:rand_comm}
\setlength{\tabcolsep}{1.05em}
\centering
\begin{tabular}{c|cc|cc|cc}
\hline
\multirow{2}{*}{Dataset} & \multicolumn{2}{c|}{Throughput} & \multicolumn{2}{c|}{Memory} & \multicolumn{2}{c}{\# Boundary Nodes} \\
 & METIS & Random & METIS & Random & METIS & Random \\ \hline
Reddit (8 partitions) & 3.1$\times$ & 5.0$\times$ & 0.47$\times$ & 0.36$\times$ & 460k & 1,016k \\
ogbn-products (10 partitions) & 3.4$\times$ & 7.3$\times$ & 0.75$\times$ & 0.31$\times$ & 1,848k & 16,797k \\
Yelp (10 partitions) & 3.1$\times$ & 5.1$\times$ & 0.83$\times$ & 0.49$\times$ & 649k & 2,026k \\
\hline
\end{tabular}
\end{table*}

\niparagraph{Generalization Improvement.}
To understand the effect of BNS-GCN's generalization capability, we also evaluate the test-accuracy convergence in Figure~\ref{fig:convergence_epoch}.
Here ogbn-products is adopted as the study case because the distribution of its test set largely differs from that of its training set \citep{hu2020open}.
From Figure~\ref{fig:convergence_epoch}, we observe that full-graph training without boundary node sampling ($p=1$) or completely isolated training ($p=0$) can overfit rapidly, regardless of different number of partitions.
With boundary node sampling ($p=0.1/0.01$), this overfitting issue is mitigated, i.e., both the convergence and the optimality are improved substantially and consistently across different number of partitions.
This is because BNS-GCN randomly modifies the graph throughout end-to-end training. 
More convergence curves on other datasets can be found in Appendix~\ref{sec:additional_convergence}.

\begin{figure}[t]
    \centering
    \includegraphics[width=1\linewidth]{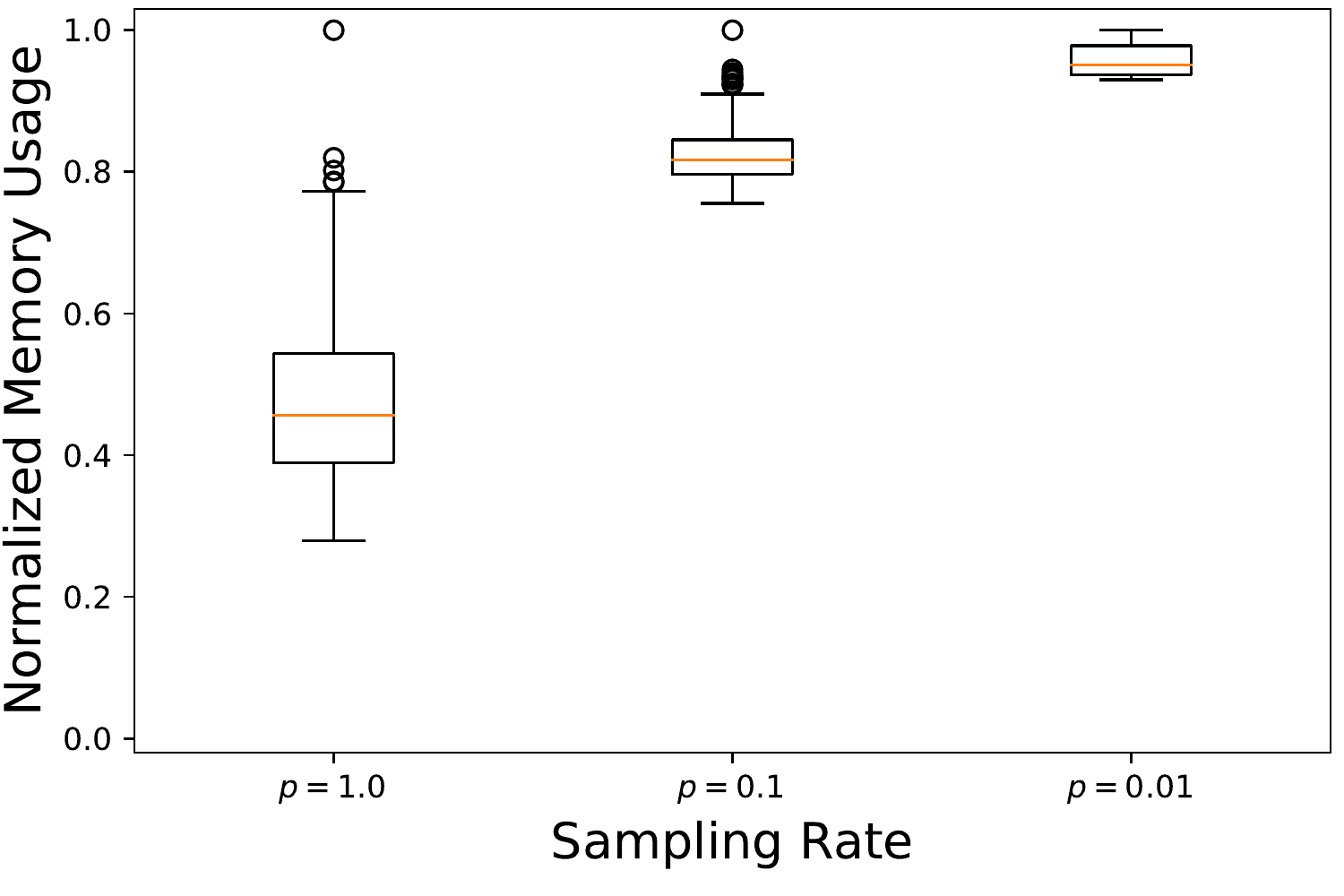}
    \caption{
    Normalized per-partition memory usage on ogbn-papers100M (192 partitions).
    The normalization is against the highest memory partition and is separated for different sampling rates $p$ of BNS-GCN.}
    \label{fig:bd_box}
\end{figure}

\niparagraph{Balanced Memory Usage.} 
To validate the benefit of BNS-GCN in balancing memory usage across partitions, we measure per-partition memory usage of ogbn-papers100M with 192 partitions and show their box plots in Figure~\ref{fig:bd_box}.
We observe that the \textit{unsampled case ($p=1.0$) suffers from a severe memory imbalance}, where one straggler increases the memory requirement by around 20\% and more than three-fourths partitions occupy less than 60\% memory. 
By contrast, with boundary node sampling, \textit{$p=0.1/0.01$ balances the memory usage} and thus better utilizes memory resource, i.e., all partitions leverage more than 70\% memory.

\begin{table*}[h]
\centering
\setlength{\tabcolsep}{1.2em}
\caption{Comparison between BNS-GCN and edge sampling methods, DropEdge and Boundary Edge Sampling (BES).}
\begin{tabular}{ccccc}
\hline
Dataset & Method & Epoch Comm (MB) & Epoch Time (sec) & Test Score (\%) \\
\hline
\multirow{3}{*}{\begin{tabular}[c]{@{}c@{}}Reddit\\(2 partitions)\end{tabular}}& DropEdge         & 301.3 & 0.613 & 97.12\\
                              & BES             & 207.9 & 0.484 & 97.16\\
                              &\textbf{BNS-GCN} & \textbf{30.4} & \textbf{0.319} & \textbf{97.17}\\
\hline
\multirow{3}{*}{\begin{tabular}[c]{@{}c@{}}ogbn-products\\(5 partitions)\end{tabular}}& DropEdge         & 1364.0 & 0.938 & \textbf{79.38}\\
                              & BES             & 521.1 & 0.551 & 79.31\\
                              &\textbf{BNS-GCN} & \textbf{138.7} & \textbf{0.388} & 79.36\\
\hline
\multirow{3}{*}{\begin{tabular}[c]{@{}c@{}}Yelp\\(3 partitions)\end{tabular}}     & DropEdge        & 718.7 & 0.606 & 65.30 \\
                              & BES             & 195.3 & 0.328 & 65.30 \\
                              &\textbf{BNS-GCN} & \textbf{75.7}  & \textbf{0.270} & \textbf{65.32} \\
\hline
\end{tabular}
\label{tab:bes}
\end{table*}

\subsection{Ablation Studies}

\niparagraph{BNS-GCN with Random Partition.}
To further understand the effectiveness of BNS-GCN and whether it relies on the adopted METIS partitioner, we conduct an ablation study by replacing METIS with random partition (i.e., randomly assign nodes to each partition without optimization) and provide the resulting accuracy in Table~\ref{tab:rand_part}.
We observe that when $p=0.1$, i.e., sampling normally, random partition plus BNS-GCN still offers a comparable performance (-0.20$\sim$+0.27) as the original METIS plus BNS-GCN, thus showing that \textit{the proposed BNS-GCN is orthogonal to the adopted graph partitioning technique} and is not necessarily limited to METIS. 
We further study whether BNS-GCN consistently improves training efficiency on top of different partition algorithms, and demonstrate the result in Table~\ref{tab:rand_comm}. 
We observe that random partition gains more benefits in throughput improvement and memory saving from BNS-GCN ($p=0.1$) than the METIS, because the former partitioner creates more boundary nodes.

\niparagraph{The Special Case $p=0$.}
The special case of the proposed boundary node sampling, $p=0$, is not recommended to use in practice.
First, $p=0$ always suffers from the worst test accuracy/score (compared with other cases ($p>0$)) on all datasets and with different partition methods (see Table~\ref{tab:acc} and Table~\ref{tab:rand_part}).
Specifically, with random partition, $p=0$ drops the test accuracy on Reddit from 97.11\% ($p=1$) to 93.37\% (lower than FastGCN~\citep{chen2018fastgcn}).
This drop is due to the absolute isolation of each partition after completely removing all boundary nodes, leading to no boundary node features during neighbor aggregation (Equation~\ref{fml:aggr}) throughout the end-to-end training. 
Second, $p=0$ also suffers from the slowest convergence regardless of different numbers of partitions or different datasets.
Third, $p=0$ overfits severely (see Figure~\ref{fig:convergence_epoch}).
Therefore, we suggest a small but non-zero sampling rate ($p=0.1/0.01$).
More discussion regarding the choice of sampling rate can be found in Appendix~\ref{sec:chocie_p}.

\niparagraph{BNS-GCN vs. Boundary Edge Sampling.}
\label{sec:compare_edge_sampling}
As many works~\citep{zhu2019aligraph,zheng2020distdgl,fey2021gnnautoscale} assume that communication overhead of distributed GCN training is caused by the inter-partition \textit{edges} (rather than boundary \textit{nodes}) and thus pursuing a minimal edge cut, one could reduce communication overhead by cutting edges using sampling techniques like DropEdge~\citep{rong2019dropedge} or even an enhanced version that samples only the boundary edges (rather than at a global scale).
To understand this, we implemented the enhanced version, dubbed \textit{\textbf{B}oundary \textbf{E}dge \textbf{S}ampling (\textbf{BES})}, and apply both BES and DropEdge to the partition-parallel training.
Table~\ref{tab:bes} compares their performance with BNS-GCN. 
For a fair comparison, all methods drop the same number of edges with BNS-GCN ($p=0.1$) over the full graph.
We observe that \textit{edge-based sampling methods are ineffective}.
For Reddit, DropEdge and BES cause 10$\times$ and 7$\times$ communication overhead of BNS-GCN, and thus 2.0$\times$ and 1.4$\times$ the overall training time.
This is because, in real-world graphs, \textit{multiple boundary edges can connect to the same boundary nodes}.
Even if we drop some of those edges, the remaining undropped edges still demand communicating the connected boundary nodes to satisfy neighbor aggregation of GCNs. 
Obviously, to eradicate such communication costs, boundary \textit{nodes} should be directly targeted and dropped, instead of using boundary \textit{edges}. 
For ogbn-products and Yelp, the advantage of BNS-GCN still holds, where BNS-GCN reduces up to 90\% communication volume and speeds up training time by up to 2.4$\times$.
Analytically, we've shown that the communication cost of distributed GCN training is only proportional to the number of boundary \textit{nodes} (see Equation~\ref{fml:comm}).

\begin{table}[t]
\setlength{\tabcolsep}{0.25em}
    \centering
    \caption{Epoch training time speedup on GAT.}
    \begin{tabular}{c|cccc}
        \hline
        BNS-GCN   & Reddit    & ogbn-products	& Yelp \\
        \hline
        $p=1$     & 1.00$\times$ (0.84s)	    & 1.00$\times$ (0.71s)         & 1.00$\times$ (0.33s) \\
        $p=0.1$	  & 1.53$\times$              & 1.78$\times$                 & 1.83$\times$ \\
        $p=0.01$  & 1.58$\times$	            & 1.91$\times$                 & 2.06$\times$ \\
        $p=0$     & 1.68$\times$              & 2.03$\times$                 & 2.20$\times$ \\
        \hline
    \end{tabular}
    \label{tab:gat}
\end{table}
\niparagraph{BNS-GCN Benefit on GAT.}
To validate the general applicability of BNS-GCN across different types of GCN models (i.e., not just GraphSAGE), we train GAT~\citep{velivckovic2017graph} with BNS-GCN and provide the improvement for a 2-layer GAT with 10 partitions in Table \ref{tab:gat}. 
We observe that BNS-GCN is consistently effective and speedups the training by 58\%$\sim$106\%, despite GAT being more computationally intensive than GraphSAGE.

%% file: section/conclusion.tex
\section{Conclusion}
While training GCNs at scale is challenging and increasingly important, 
existing methods for distributed GCN training are still limited in their achievable performance and scalability. 
This work takes the initial effort to analyze the three major challenges in distributed GCN training and then identify their underlying cause. On top of that, we propose an efficient and scalable method  for full-graph GCN training, BNS-GCN, and then validate its effectiveness through both theoretical analysis and extensive empirical evaluations.  
We believe that these findings and the proposed method have provided a better understanding of distributed GCN training and will inspire further innovations in this direction.

%% file: section/appendix.tex
\onecolumn

\section{The Variance Analysis}
\label{sec:var_proof}

In this section, we derive the variance of embedding approximation when using our proposed BNS-GCN method, of which the result is listed in Table~\ref{tab:variance} of the main content. 

For a given graph $\mathcal{G}=(\mathcal{V},\mathcal{E})$ with an adjacency matrix $A$, we define the propagation matrix $P$ as $P=\tilde{D}^{-1/2}\tilde{A}\tilde{D}^{-1/2}$, where $\tilde{A}=A+I,\tilde{D}_{u,u}=\sum_v\tilde{A}_{u,v}$. One GCN layer performs one step of feature propagation \citep{kipf2016semi} as formulated below:

\begin{align}
Z^{(\ell)}&=PH^{(\ell-1)}W^{(\ell-1)} \label{eq:gcn_step1}\\
H^{(\ell)}&=\sigma\left(Z^{(\ell)}\right)\notag
\end{align}

where $H^{(\ell)}$, $W^{(\ell)}$, and $Z^{(\ell)}$ denote the embedding matrix, the trainable weight matrix, and the intermediate embedding matrix in the $\ell$-th layer, respectively, and $\sigma$ denotes the non-linear function. 
Without loss of generality, we provide our analysis for one layer of GCNs and drop the superscripts of $(\ell)$ and $(\ell-1)$ in the reminder of the discussion for simplicity.

For distributed GCN training using partition-parallelism, if denoting the inner node set and the boundary node set of the $i$-th partition as $\mathcal{V}_i$ and $\mathcal{B}_i$, respectively, the operations of the $i$-th partition
for calculating Equation~\ref{eq:gcn_step1} are as follows:
$$Z_{\mathcal{V}_i,*}=
\begin{bmatrix}
P_{\mathcal{V}_i,\mathcal{V}_i} & P_{\mathcal{V}_i,\mathcal{B}_i}
\end{bmatrix}
\begin{bmatrix}
H_{\mathcal{V}_i,*} \\
H_{\mathcal{B}_i,*}
\end{bmatrix}W$$

In BNS-GCN, $Z_{\mathcal{V}_i,*}$ is approximated as $\tilde{Z}_{\mathcal{V}_i,*}$ due to its boundary node sampling, i.e.,  

$$\tilde{Z}_{\mathcal{V}_i,*}=
\begin{bmatrix}
P_{\mathcal{V}_i,\mathcal{V}_i} & P_{\mathcal{V}_i,\mathcal{U}_i}
\end{bmatrix}
S
\begin{bmatrix}
H_{\mathcal{V}_i,*}\\
H_{\mathcal{U}_i,*}
\end{bmatrix}W$$
where $\mathcal{U}_i$ denotes the sampled boundary node set, $p$ denotes the sampling rate, and $S$ is a diagnal matrix with its elements being defined as: 

\begin{align}
S_{u,u}=\begin{cases}
    1,     & u\leq|\mathcal{V}_i|\notag\\
    1/p,   & u>|\mathcal{V}_i|
\end{cases}
\end{align}

Similar to the variance analysis in \citep{chen2018stochastic} and \citep{zou2019layer}, our goal is to compute the average variance of the approximated embedding for one GCN layer, which can be defined as $\expe_{\mathcal{U}}[\|\tilde{Z}-Z\|_F^2]/|\mathcal{V}|$. In our analysis, we adopt the same assumption as that in \citep{zou2019layer}, which bounds the matrix product $HW$ as follows:

\begin{assumption}
\label{asm:hw}
We assume that the $L_2$-norm of each row in HW is upper bounded by a constant, i.e.,
there exists a constant $\gamma$ such that $\|H_{u,*}W\|_2 \leq\gamma$ for all $u\in \left|\mathcal{V}\right|$.
\end{assumption}

Next, we calculate the total variance of the embedding approximation for the $i$-th partition:

\allowdisplaybreaks

\begin{align}
\expe_{\mathcal{U}_i}[\|\tilde{Z}_{\mathcal{V}_i,*}-Z_{\mathcal{V}_i,*}\|_F^2]
	& =\expe_{\mathcal{U}_i}\left[\left\|
\begin{bmatrix}
P_{\mathcal{V}_i,\mathcal{V}_i} & P_{\mathcal{V}_i,\mathcal{U}_i}
\end{bmatrix}
S
\begin{bmatrix}
H_{\mathcal{V}_i,*}\\
H_{\mathcal{U}_i,*}
\end{bmatrix}W
-
\begin{bmatrix}
P_{\mathcal{V}_i,\mathcal{V}_i} & P_{\mathcal{V}_i,\mathcal{B}_i}
\end{bmatrix}
\begin{bmatrix}
H_{\mathcal{V}_i,*} \\
H_{\mathcal{B}_i,*}
\end{bmatrix}W
		\right\|_F^2\right]  \notag\\
	& =\expe_{\mathcal{U}_i}\left[\left\|\frac{1}{p}P_{\mathcal{V}_i,\mathcal{U}_i}H_{\mathcal{U}_i,*}W-P_{\mathcal{V}_i,\mathcal{B}_i}H_{\mathcal{B}_i,*}W\right\|_F^2\right]  \label{eq:step2}\\
	& =\sum_{v\in\mathcal{V}_i}\expe_{\mathcal{U}_i}\left[\left\|\sum_{u\in\mathcal{U}_i}\frac{1}{p}P_{v,u}H_{u,*}W-P_{v,\mathcal{B}_i}H_{\mathcal{B}_i,*}W\right\|_2^2\right]  \notag\\
	& =\sum_{v\in\mathcal{V}_i}\frac{1}{p}\sum_{u\in\mathcal{B}_i}\left\|P_{v,u}H_{u,*}W-\frac{1}{|\mathcal{B}_i|}P_{v,\mathcal{B}_i}H_{\mathcal{B}_i,*}W\right\|_2^2 \label{eq:step4}\\
	& =\sum_{v\in\mathcal{V}_i}\frac{1}{p}\left(\sum_{u\in\mathcal{B}_i}\left\|P_{v,u}H_{u,*}W\right\|_2^2-\left\|P_{v,\mathcal{B}_i}H_{\mathcal{B}_i,*}W\right\|_2^2\right) \notag\\
	& \leq\frac{1}{p}\sum_{v\in\mathcal{V}_i}\sum_{u\in\mathcal{B}_i}\left\|P_{v,u}H_{u,*}W\right\|_2^2 \notag
\end{align}

where the step of Equation \ref{eq:step2} removes the common factor $P_{\mathcal{V}_i,\mathcal{V}_i}H_{\mathcal{V}_i,*}W$ and the step of Equation \ref{eq:step4} uses the fact that the selection of nodes in $\mathcal{B}_i$ are independent.

Based on Assumption \ref{asm:hw}, we have $\left\|H_{u,*}W\right\|_2\leq\gamma$. As a result, the above upper bound can be further written as: 

\begin{align*}
\expe_{\mathcal{U}_i}[\|\tilde{Z}_{\mathcal{V}_i,*}-Z_{\mathcal{V}_i,*}\|_F^2]
	& \leq\frac{1}{p}\sum_{v\in\mathcal{V}_i}\sum_{u\in\mathcal{B}_i}P_{v,u}^2\gamma^2 \\
	&=\frac{1}{p}\gamma^2\left\|P_{\mathcal{V}_i,\mathcal{B}_i}\right\|_F^2
\end{align*}

Thus, \textit{\textbf{the total variance of the embedding approximation for the $i$-th partition is $\mathcal{O}(|\mathcal{B}_i|\gamma^2/s_{\ell})$}} as shown in Table~\ref{tab:variance} of the main content, where $s_{\ell}$ denote the size of the sampled node set.

Finally, the global average variance can be calulated as:
\begin{align*}
\frac{\expe_{\mathcal{U}}[\|\tilde{Z}-Z\|_F^2]}{|\mathcal{V}|}
	&=\frac{\sum_{i}\expe_{\mathcal{U}_i}[\|\tilde{Z}_{\mathcal{V}_i,*}-Z_{\mathcal{V}_i,*}\|_F^2]}{|\mathcal{V}|} \\
	&\leq\frac{\gamma^2\sum_{i}\left\|P_{\mathcal{V}_i,\mathcal{B}_i}\right\|_F^2}{p|\mathcal{V}|} \\
	&\leq\frac{\gamma^2\|P\|_F^2}{p|\mathcal{V}|}
\end{align*}

\clearpage

\section{Convergence Speedup (Additional Experiments)}
\label{sec:additional_convergence}
\begin{figure}[h]
  \centering
  \includegraphics[width=1\linewidth]{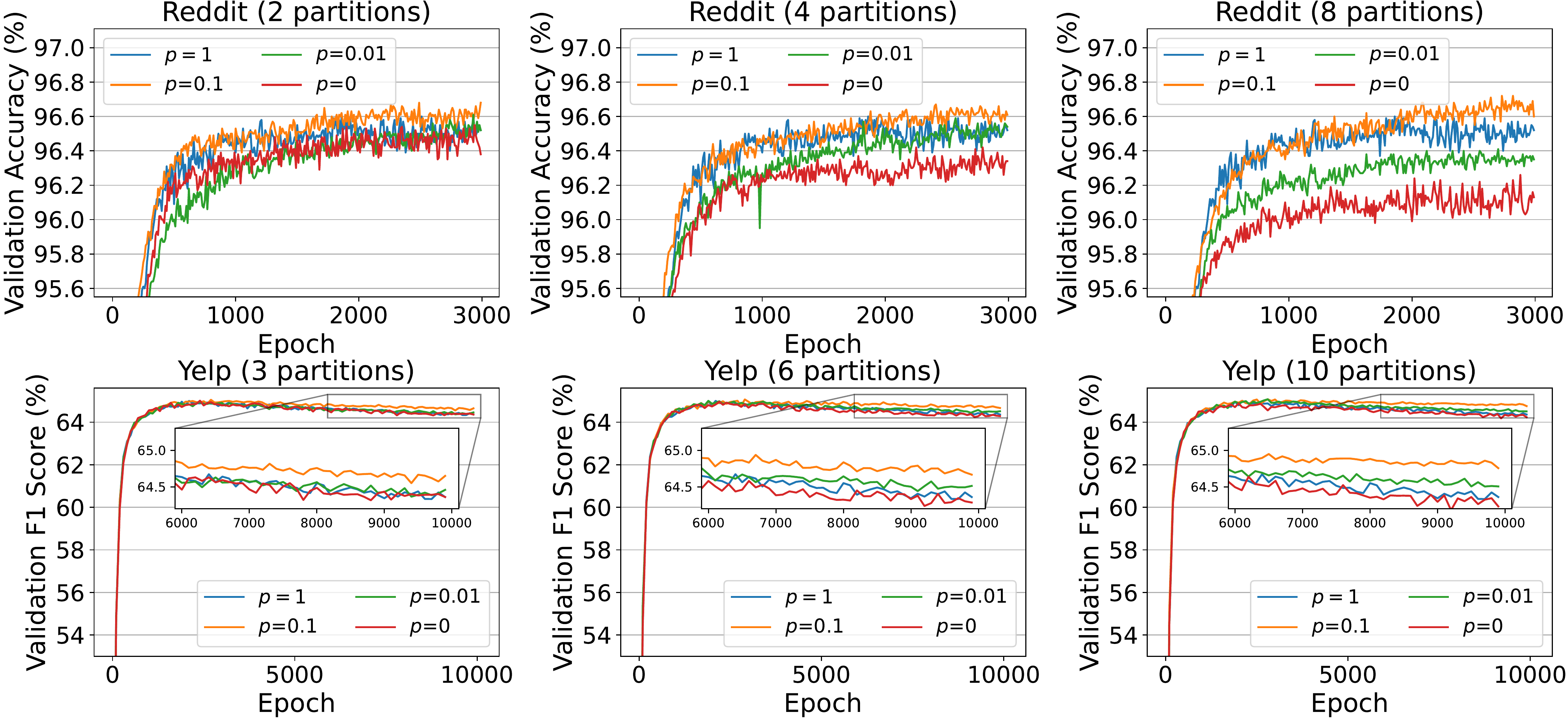}
  \caption{Convergence comparison between unsampled full-graph training (BNS-GCN with $p=1$), boundary-node sampled training ($0<p<1$), and isolated training ($p=0$) on Reddit and Yelp.}
  \label{fig:convergence_epoch_2}
\end{figure}

Figure~\ref{fig:convergence_epoch_2} shows convergence speedups of BNS-GCN on more datasets under the same setting of Figure~\ref{fig:convergence_epoch} of the main content.
\textit{\textbf{The observations from the main content still hold.}}
Especially, \textit{boundary node sampling at a high rate ($p=0.1$) achieves the best convergence} regardless of different numbers of partitions or different datasets.
A lower rate ($p=0.01$) still remains a close convergence as $p=0.1$.
The special case \textit{$p=0$ suffers from not only the worst convergence but also increased convergence gap} between $p=0$ and $p=0.1$ as more partitions are involved, because of complete removal of boundary node information.
Lastly, $p=1$ and $p=0$ can still overfit (see Yelp) but \textit{boundary node sampling ($p=0.1/0.01$) mitigates the overfitting} by random modification of the graph throughout training.

\clearpage

\section{Efficiency Comparison with Sampling-based Methods}
\label{sec:sampling_throughput}
Table~\ref{tab:speed_over_sampling} compares the training efficiency between the popular sampling-based methods and BNS-GCN under the same settings of the main content.
As can be seen, \textit{\textbf{BNS-GCN outperforms the sampling-based methods with a great margin, while achieving a higher accuracy}} (see Table~\ref{tab:acc} of the main content).
\begin{table}[h]
\centering
\caption{Comparison of training efficiency on Reddit, where BNS-GCN with various \textit{boundary sampling rates} under 8 partitions are shown.}
\label{tab:speed_over_sampling}
\setlength{\tabcolsep}{0.4em}
\begin{tabular}{ll|ccccccc}
\hline
& Method                                                              & GraphSAGE & FastGCN & VR-GCN & ClusterGCN & \textbf{BNS-GCN(1)} & \textbf{BNS-GCN(0.1)} & \textbf{BNS-GCN(0.01)} \\ \hline
& \begin{tabular}[c]{@{}l@{}}Train time \\ per epoch \end{tabular} &       6.20s    & 5.08s     & 3.85s  &  1.35s & 0.777s & 0.198s & 0.150s \\ \hline
& Speedup                                                             & 1$\times$        &   1.22$\times$   &  1.61$\times$ &  4.59$\times$ & 8.0$\times$ & \textbf{31.3$\times$} & \textbf{41.3$\times$} \\ 
\hline
\end{tabular}
\end{table}

\clearpage

\section{Overhead of Boundary Node Sampling}
\label{sec:sampling_overhead}
In this section, we evaluate the overhead introduced by the \textit{boundary node sampling} of BNS-GCN under different sampling rates and number of partitions, and also compare it with the overhead of the state-of-the-art sampling methods.
Table~\ref{tab:overhead} summarizes the results.
We observe that \textit{\textbf{node, edge, and random walk sampling can introduce a non-trivial overhead, which is up to 24\% of training time}}~\citep{zeng2019graphsaint}.
By contrast, \textbf{\textit{boundary node sampling incurs only a negligible overhead, i.e., 0\%$\sim$7.3\%}}, because it only needs to perform sampling on the boundary region instead of the whole graph as used in the state-of-the-art methods.
Also, the light weightiness of boundary nodes sampling lies in its parallizability across partitions, instead of requiring sequential processing.
Besides, we also compare BNS-GCN with the graph-level sampling method such as ClusterGCN~\citep{chiang2019cluster}.
We find that the overhead of boundary node sampling is still much lower than ClusterGCN, because ClusterGCN needs to merge multiple subgraphs into one cluster with a sampling time roughly proportional to the number of edges in the whole graph. 
By contrast, boundary node sampling only needs to modify those boundary edges of selected boundary nodes, and its sampling time is proportional only to the number of boundary edges, which is only a fraction of ClusterGCN.

\begin{table}[h]
\centering
\caption{Comparison of BNS-GCN's sampling overhead with the state-of-the-art methods in GraphSAINT~\citep{zeng2019graphsaint} on Reddit, where the overhead percentage is calculated by the sampling time divided by the training time.}
\label{tab:overhead}
\begin{tabular}{cccc}
\hline
\multicolumn{4}{c}{The state-of-the-art samplers}                                     \\ \hline
\multicolumn{1}{c|}{Node}          & \multicolumn{3}{c}{23\%}                         \\
\multicolumn{1}{c|}{Edge}          & \multicolumn{3}{c}{20\%}                         \\
\multicolumn{1}{c|}{Random walk}   & \multicolumn{3}{c}{24\%}                         \\ \hline
\multicolumn{4}{c}{\textbf{BNS-GCN sampler}}                                          \\ \hline
\multicolumn{1}{c|}{\# Partitions} & 2              & 4              & 8              \\ \hline
\multicolumn{1}{c|}{$p=1.00$}         & \multicolumn{3}{c}{\textbf{0\%}}                 \\
\multicolumn{1}{c|}{$p=0.10$}       & \textbf{1.7\%} & \textbf{3.2\%} & \textbf{6.6\%} \\
\multicolumn{1}{c|}{$p=0.01$}      & \textbf{1.3\%} & \textbf{3.0\%} & \textbf{7.3\%} \\
\multicolumn{1}{c|}{$p=0.00$}         & \multicolumn{3}{c}{\textbf{0\%}}                 \\
\hline
\end{tabular}
\end{table}

\clearpage

\section{The Choice of $p$}
\label{sec:chocie_p}
In this section, we discuss how to choose the \textit{boundary node sampling rate} $p$ in practice for maximizing the efficiency of GCN training.
Empirically, \textit{$p=0.1$ combines the best of all worlds}: throughput boosting, communication reduction, memory saving, convergence speedup, and final accuracy, as well as sampling overhead, across different number of partitions and different datasets, according to our extensive experiments.
To further validate this, we compare the test accuracy of $p$ between 0.1 and 1 and summarize the results in Table~\ref{tab:choice_of_p}.
We can see that the advantage of $p=0.1$ still holds, i.e., offering similar accuracy but less communication/memory compared with higher $p$ values.

\begin{table}[h]
\centering
\caption{Test accuracy of BNS-GCN with a sampling rate $p$ between 0.1 and 1.}
\label{tab:choice_of_p}
\begin{tabular}{c|ccccc}
\hline
Dataset & $p=0.1$ & $p=0.3$ & $p=0.5$ & $p=0.8$ & $p=1.0$ \\
\hline
Reddit (2 partitions) & 97.17\% & 97.18\% & 97.15\% & 97.13\%  & 97.11\%\\
\hline
ogbn-product (5 partitions) & 79.36\% & 79.30\% & 79.34\% & 79.24\% & 79.14\%\\
\hline
\end{tabular}
\end{table}

\clearpage
\section{Artifact Appendix} 
\subsection{Abstract}
Our artifact contains the full source code of BNS-GCN. 
It includes both the baseline (vanilla partition-parallel training) and the proposed boundary-node-sampled training for GCNs on various datasets.
Running the code requires a machine (at least 120 GB host memory) with multiple (at least five) Nvidia GPUs (at least 11GB each).
Software is provided with our docker image.
With the aforementioned hardware and software, running our provided scripts will validate the main experiments in the paper, such as per-epoch training time, training time breakdown, memory usage, and accuracy.

\subsection{Artifact check-list (meta-information)}

{\small
\begin{itemize}
  \item {\bf Algorithm: }Graph Convolutional Network (GCN), Distributed Training, Random Sampling
  \item {\bf Data set: }Reddit, ogbn-products, Yelp (all included in our docker or software setup)
  \item {\bf Run-time environment: }Ubuntu 18.04, Python 3.8, CUDA 11.1, PyTorch 1.8.0, DGL 0.7.0, OGB 1.3.0
  \item {\bf Hardware: }A X64-CPU machine with at least 120 GB host memory, at least five Nvidia GPUs (at least 11GB each).
  \item {\bf Execution: }Bash scripts, Running each experiment takes less than 1 hour.
  \item {\bf Metrics: }Training time, training time breakdown, memory usage, accuracies
  \item {\bf Output: }Console, and log file
  \item {\bf How much disk space required (approximately)?: } 50GB
  \item {\bf How much time is needed to prepare workflow (approximately)?: }30 minutes
  \item {\bf How much time is needed to complete experiments (approximately)?: }10 hours
  \item {\bf Publicly available?: }yes
  \item {\bf Code licenses (if publicly available)?: } MIT License
  \item {\bf Archived (provide DOI)?: } 10.5281/zenodo.6079700
\end{itemize}
}
\subsection{Description}

\subsubsection{How delivered}
\begin{itemize}
    \item Source code in the archival repository for ACM badges: \href{https://doi.org/10.5281/zenodo.6079700}{https://doi.org/10.5281/zenodo.6079700}.
    \item Latest source code in GitHub repository: \href{https://github.com/RICE-EIC/BNS-GCN}{https://github.com/RICE-EIC/BNS-GCN}.
    \item Docker image: \href{https://hub.docker.com/r/cheng1016/bns-gcn}{https://hub.docker.com/r/cheng1016/bns-gcn}.
    \item Approximate disk space: 50GB, used for large datasets
\end{itemize}

\subsubsection{Hardware dependencies}
\begin{itemize}
    \item A X86-CPU machine with at least 120 GB host memory 
    \item At least five Nvidia GPUs (at least 11 GB each)
\end{itemize}

\subsubsection{Software dependencies}
All provided in our docker image:
\begin{itemize}
    \item Ubuntu 18.04
    \item Python 3.8
    \item CUDA 11.1
    \item PyTorch 1.8.0
    \item customized DGL 0.7.0
    \item OGB 1.3.0
\end{itemize}

\subsubsection{Data sets}
All dataset (Reddit, ogbn-products, Yelp) are either included in our docker image or to be downloaded by our scripts.

\subsection{Installation}
Detailed instructions are provided in \texttt{README.md} in our GitHub repository. 
For example, just run \texttt{docker pull cheng1016/bns-gcn} followed by \texttt{docker run -{}-gpus all -it cheng1016/bns-gcn}.

\subsection{Experiment workflow}
The workflow involves invoking top-level \texttt{main.py} which then drives other modules for distributed GCN training.
All ``one-click-to-run'' scripts to reproduce main experiments in the paper are provided in the \texttt{scripts/*.sh} in our GitHub repository.

\subsection{Evaluation and expected result}
All steps are in \texttt{scripts/*.sh} in our GitHub repository.

\subsection{Experiment customization}
We provide a detailed guide for customization in \texttt{README.md} in our GitHub repository.
Hyper-parameters and configurations can be customized by the options fed to \texttt{main.py}, e.g., allowing users to choose the number of training epochs, the number of graph partitions (or GPUs), different partitioning methods, and even extending training to multiple machines with multiple GPUs.

\subsection{Notes}
For the hyper-scale dataset ogbn-papers100M, the experiment was conducted on 32 machines, each of which has 6 Tesla V100 (16GB) with IBM Power9 (605GB).